\documentclass{article}

\usepackage[final, nonatbib]{neurips_2021}

\usepackage[utf8]{inputenc} 
\usepackage[T1]{fontenc}    
\usepackage{hyperref}       
\usepackage{url}            
\usepackage{booktabs}       
\usepackage{amsfonts}       
\usepackage{nicefrac}       
\usepackage{microtype}      
\usepackage{xcolor}

\usepackage{graphicx}
\usepackage{chngcntr}
\usepackage{xcolor}
\usepackage{listings}
\usepackage{subfigure}
\usepackage{makecell}
\usepackage{stmaryrd}
\usepackage{wrapfig}
\def\c{\char95}

\title{Modeling Heterogeneous Hierarchies with Relation-specific Hyperbolic Cones}

\author{%
   Yushi Bai\thanks{Equal contribution} \\
   \texttt{bys18@mails.tsinghua.edu.cn} \\
   Tsinghua University\\
    \And
    Rex Ying$^*$ \\
    \texttt{rexying@stanford.edu} \\
    Stanford University \\
    \And
    Hongyu Ren \\
    \texttt{hyren@cs.stanford.edu} \\
    Stanford University \\
    \And
    Jure Leskovec \\
    \texttt{jure@cs.stanford.edu} \\
    Stanford University \\
}
\usepackage{booktabs}
\usepackage{amsfonts}
\usepackage{microtype}
\usepackage{url}
\usepackage{verbatim}
\usepackage{graphicx}
\usepackage{caption} 
\usepackage{multirow}
\usepackage{xspace}
\usepackage{epsfig}
\usepackage{amsmath}
\usepackage{amsthm}
\usepackage{amssymb}
\usepackage{times}
\usepackage{xr}
\usepackage{bbm}
\usepackage{dsfont}
\usepackage{bm}
\usepackage{enumitem}
\usepackage{hyperref}

\newcommand{\xhdr}[1]{{\noindent\bfseries #1}.}

\newcommand{\name}{ConE\xspace}

\newcommand{\cut}[1]{}

\newtheorem{theorem}{Theorem}

\newtheorem{definition}{Definition}

\newcommand{\xproof}{\noindent {\it Proof. }}

\newenvironment{itemize*}%
  {\begin{itemize}[topsep=0pt, partopsep=0pt]%
    \setlength{\itemsep}{0pt}%
    \setlength{\parskip}{0pt}}%
  {\end{itemize}}
  
\newenvironment{enumerate*}%
  {\begin{enumerate}[topsep=-5pt, partopsep=0pt]%
    \setlength{\itemsep}{0pt}%
    \setlength{\parskip}{0pt}%
    \setlength{\parsep}{0pt}}%
  {\end{enumerate}}

\definecolor{codegreen}{rgb}{0,0.4,0}
\definecolor{codegray}{rgb}{0.4,0.4,0.4}
\definecolor{codepurple}{rgb}{0.5,0,0.7}
\definecolor{backcolour}{rgb}{0.96,0.96,0.94}
 
\lstdefinestyle{mystyle}{
    backgroundcolor=\color{backcolour},   
    commentstyle=\color{codegreen},
    keywordstyle=\color{magenta},
    numberstyle=\tiny\color{codegray},
    stringstyle=\color{codepurple},
    basicstyle=\ttfamily\footnotesize,
    breakatwhitespace=false,         
    breaklines=true,                 
    captionpos=b,                    
    keepspaces=true,                 
    numbers=left,                    
    numbersep=5pt,                  
    showspaces=false,                
    showstringspaces=false,
    showtabs=false,                  
    tabsize=2
}
 
\begin{document}

\maketitle

\begin{abstract}
Hierarchical relations are prevalent and indispensable for organizing human knowledge captured by a knowledge graph (KG).
The key property of hierarchical relations is that they induce a partial ordering over the entities, which needs to be modeled in order to allow for hierarchical reasoning. 
However, current KG embeddings can model only a single global hierarchy (single global partial ordering) and fail to model multiple heterogeneous hierarchies that exist in a single KG.
Here we present ConE (Cone Embedding), a KG embedding model that is able to simultaneously model multiple hierarchical as well as non-hierarchical relations in a knowledge graph. 
ConE embeds entities into hyperbolic cones and models relations as transformations between the cones.
In particular, ConE uses cone containment constraints in different subspaces of the hyperbolic embedding space to capture multiple heterogeneous hierarchies.
Experiments on standard knowledge graph benchmarks show that ConE obtains state-of-the-art performance on hierarchical reasoning tasks as well as knowledge graph completion task on hierarchical graphs.
In particular, our approach yields new state-of-the-art Hits@1 of 45.3\% on WN18RR and 16.1\% on DDB14 (0.231 MRR). As for hierarchical reasoning task, our approach outperforms previous best results by an average of 20\% across the three datasets.
\end{abstract}

\section{Introduction}

Knowledge graph (KG) is a data structure that stores factual knowledge in the form of triplets, which connect two entities (nodes) with a relation (edge) \cite{hogan2021knowledge}.
Knowledge graphs play an important role in many scientific and machine learning applications, including question answering \cite{hao2017end}, information retrieval \cite{xiong2017explicit} and discovery in biomedicine \cite{zitnik2018modeling}.
Knowledge graph completion is the problem of predicting missing relations in the graph, and is crucial in many real-world applications.
Knowledge graph embedding (KGE) models \cite{bordes2013translating,lin2015learning,sun2019rotate} approach the task by embedding entities and relations into low-dimensional vector space and then use the embeddings to learn a function that given a head entity $h$ and a relation $r$ predicts the tail entity $t$. 

Hierarchical information is ubiquitous in real-world KGs, such as WordNet \cite{miller1995wordnet} or Gene Ontology \cite{michael2000gene}, since much human knowledge is organized hierarchically. KGs can be composed of a mixture of non-hierarchical (e.g., \textit{likes}, \textit{friendOf}) and hierarchical (e.g., \textit{isA}, \textit{partOf}), where non-hierarchical relations capture interactions between the entities at the same level while hierarchical relations induce a tree-like partial ordering structure of entities.

\begin{figure}[t]
\centering
\subfigure[Multiple heterogeneous hierarchies in knowledge graph.]{
\begin{minipage}{0.49\linewidth}
\centering
\includegraphics[width=1\linewidth, trim=0 0 0 0,clip]{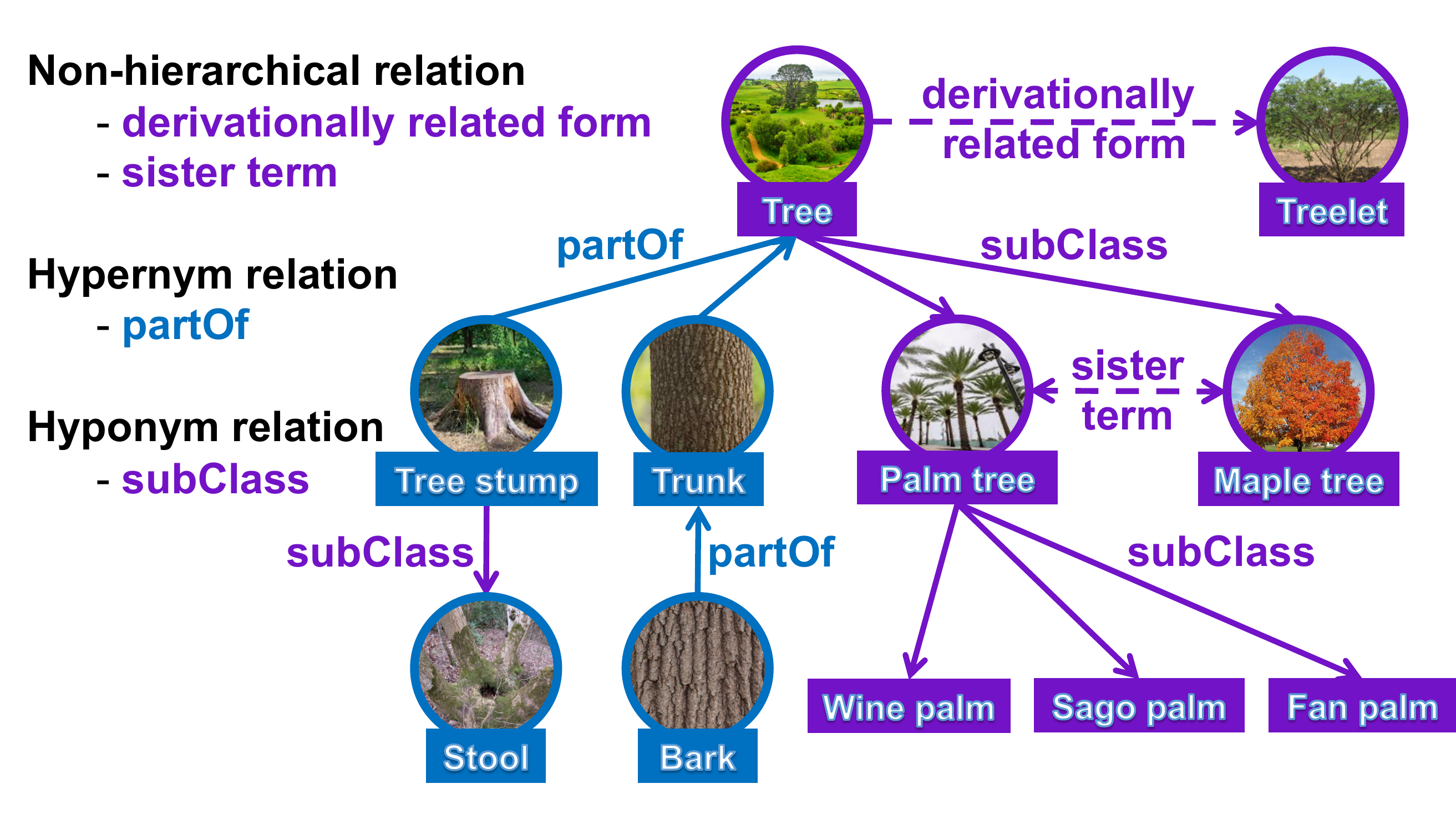}
\label{fig:hete_hier}
\end{minipage}%
}\qquad
\subfigure[Hyperbolic entailment cones in 2D hyperbolic plane for $K=0.1$.]{
\begin{minipage}{0.43\linewidth}
\centering
\includegraphics[width=0.82\linewidth,trim=40 0 40 0,clip]{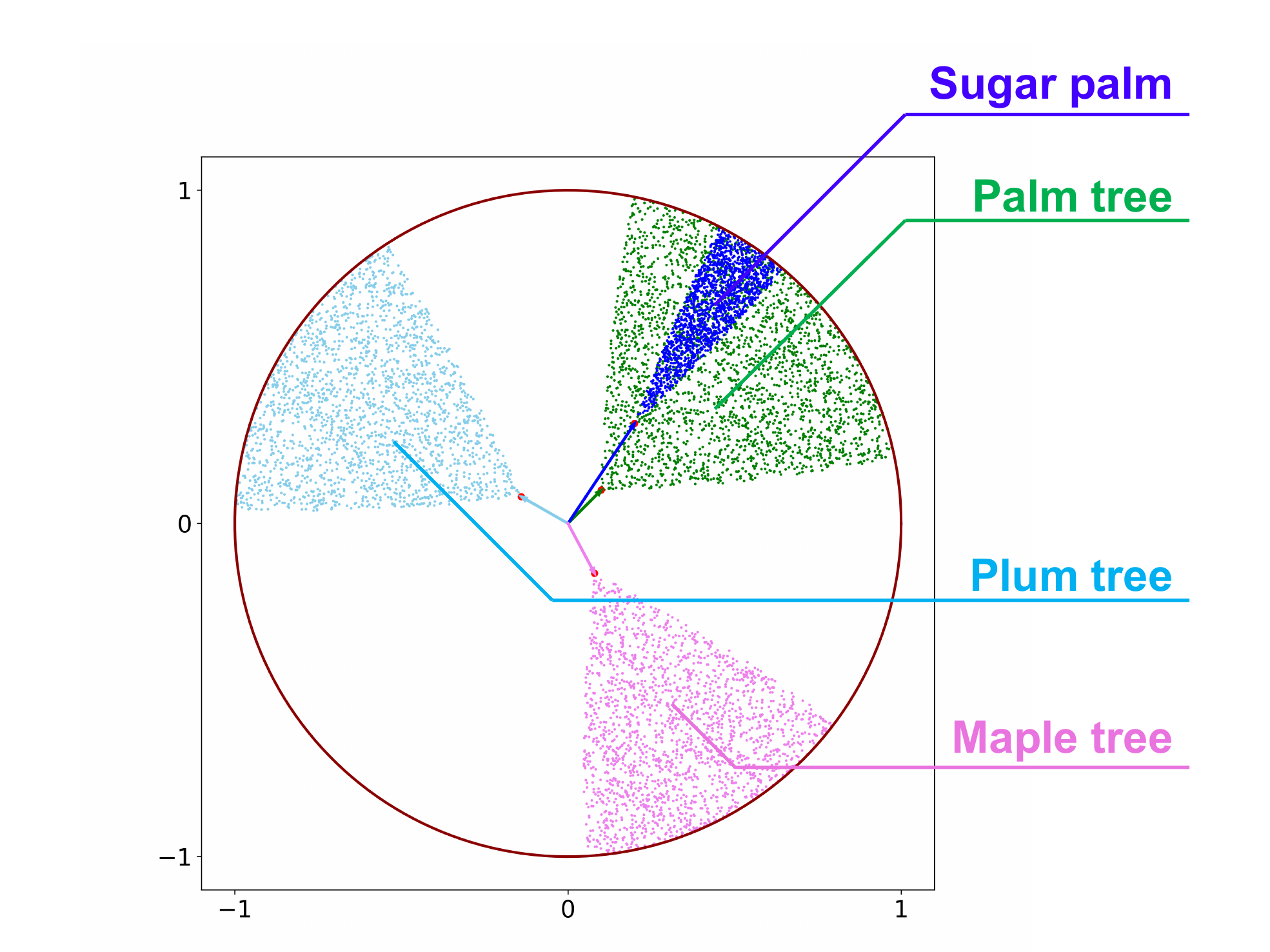}
\label{fig:cone}
\end{minipage}%
}%
\centering
\caption{(a) There are three categories of relations: non-hierarchical relation (\textit{sister term}), hypernym (\textit{partOf}) and hyponym relation (\textit{subClass}). Relations induce multiple independent hierarchies.
(b) \name uses $d$ 2D hyperbolic entailment cones to model an entity. Entities \textit{PalmTree} and \textit{SugarPalm} are connected by a hyponym relation \textit{subClass} and therefore the cone of \textit{PalmTree} contains the cone of \textit{SugarPalm}.}
\vspace{-2mm}
\end{figure}

While non-hierarchical relations have been successfully modeled in the past, there has been a recent focus on modeling hierarchical relations.
Recent works in this area propose the use of a variety of embedding geometries such as hyperbolic embeddings, box embeddings, and cone embeddings \cite{nickel2017poincare,vilnis2018probabilistic,ganea2018hyperbolic} to model partial ordering property of hierarchical relations, but two important challenges remain:
\textbf{(1)} Existing works that consider hierarchical relations \cite{nickel2018learning} do not take into account existing non-hierarchical relations \cite{balavzevic2019tucker}.
\textbf{(2)} These methods can only be applied to graphs with a single hierarchical relation type, and are thus not suitable to real-world knowledge graphs that simultaneously encode multiple hierarchies using many different relations. For example, in Figure 1, \textit{subClass} and \textit{partOf} each define a unique hierarchy over the same set of entities. However, existing models treat all relations in a KG as part of one single hierarchy, limiting the ability to reason with different types of heterogeneous hierarchical relations.
While there are methods for reasoning over KGs that use hyperbolic space (MuRP \cite{balazevic2019multi}, RotH \cite{chami2020low}), which is suitable for modeling tree-like graphs, the choice of relational transformations used in these works (rotation) prevents them from faithfully capturing all the properties of hierarchical relations. For example, they cannot model transitivity of hierarchical relations: if there exist relations $(h_1, r, h_2)$ and $(h_2, r, h_3)$, then $(h_1, r, h_3)$ exists, \emph{i.e.} $h_1$ and $h_3$ are also related by relation $r$.

Here we propose a novel hyperbolic knowledge graph embedding model \name. \name is motivated by the transitivity of nested angular cones \cite{ganea2018hyperbolic} that naturally model the partial ordering defined by hierarchical relations. Our proposed approach embeds entities into the product space of hyperbolic planes, where the coordinate in each hyperbolic plane corresponds to a 2D hyperbolic cone. To address challenge \textbf{(1)}, we model non-hierarchical relations as hyperbolic cone rotations from head entity to tail entity, while we model hierarchical relations as a restricted rotation which guarantees cone containment (Figure~\ref{fig:cone}). To address challenge \textbf{(2)}, we assign distinct embedding subspaces corresponding to product spaces of a different set of hyperbolic planes for each hierarchical relation, to enforce cone containment constraints. By doing so, multiple heterogeneous hierarchies are preserved simultaneously in unique subspaces, allowing \name to perform multiple hierarchical reasoning tasks accurately. 

We evaluate the performance of \name on the KG completion task and hierarchical reasoning task. A single trained \name model can achieve remarkable performance on both tasks simultaneously. On KG completion task, \name achieves new state-of-the-art results on two benchmark knowledge graph datasets including WN18RR \cite{bordes2013translating,dettmers2018convolutional}, DDB14 \cite{wang2020entity} (outperforming by \textbf{0.9\%} and \textbf{4.5\%} on Hits@1 metric). We also develop a novel biological knowledge graph GO21 from biomedical domain and show that \name successfully models multiple hierarchies induced by different biological processes. We also evaluate our model against previous hierarchical modeling approaches on ancestor-descendant prediction task. Results show that \name significantly outperforms baseline models (by \textbf{20\%} on average when missing links are included), suggesting that it effectively models multiple heterogeneous hierarchies. 
Moreover, \name performs well on the lowest common ancestor (LCA) prediction task, improving over previous methods by \textbf{100\%} in Hits@3 metric.

\section{Related Work}
\label{sec:related}
\xhdr{Hierarchical reasoning} The most related line of work is learning structured embeddings to perform hierarchical reasoning on graphs and ontologies: order embedding, probabilistic order embedding, box embedding, Gumbel-box embedding and hyperbolic embedding \cite{nickel2017poincare, vilnis2018probabilistic, ganea2018hyperbolic, vendrov2015order, lai2017learning, li2018smoothing, dasgupta2020improving}.
These embedding-based methods map entities to various geometric representations that can capture the transitivity and entailment of hierarchical relations.
These methods aim to perform hierarchical reasoning (transitive closure completion), such as predicting if an entity is an ancestor of another entity.
However, the limitation of the above works is that they can only model a single hierarchical relation, and it remains unexplored how to extend them to multiple hierarchical relations in heterogeneous knowledge graphs. Recently, \cite{dasgupta2021boxtobox} builds upon the box embedding and further models joint (two) hierarchies using two boxes as entity embeddings. However, the method is not scalable since the model needs to learn a quadratic number of transformation functions between all pairs of hierarchical relations. Furthermore, the missing part is that these methods do not leverage non-hierarchical relations to further improve the hierarchy modeling. For example in Figure~\ref{fig:hete_hier}, with the \textit{sisterTerm(PalmTree, MapleTree)} and \textit{subClass(PalmTree, Tree)}, we may infer \textit{subClass(MapleTree, Tree)}. In contrast to prior methods, \name is able to achieve exactly this type of reasoning as it can simultaneously model multiple hierarchical as well as non-hierarchical relations.

\xhdr{Knowledge graph embedding} Various embedding methods have been proposed to model entities and relations in heterogeneous knowledge graphs. Prominent examples include TransE \cite{bordes2013translating}, DistMult \cite{yang2014embedding}, ComplEx \cite{trouillon2016complex}, RotatE \cite{sun2019rotate} and TuckER \cite{balavzevic2019tucker}. These methods often require high embedding dimensionality to model all the triples. 
Recently KG embeddings based on hyperbolic space have shown success in modeling hierarchical knowledge graphs. 
MuRP \cite{balazevic2019multi} learns relation-specific parameters in the Poincar\'e ball model.
RotH \cite{chami2020low} uses rotation and reflection transformation in $n$-dimensional Poincar\'e space to model relational patterns, and achieves state-of-the-art for the KG completion task, especially under low-dimensionality. However, transformations used in MuRP and RotH cannot capture transitive relations which hierarchical relations naturally are.

To the best of our knowledge, \name is the first model that can faithfully model multiple hierarchical as well as non-hierarchical relations in a single embedding framework.

\section{\name Model Framework}
\label{sec:model}
\subsection{Preliminaries}
\xhdr{Knowledge graphs and knowledge graph embeddings}
We denote the entity set and the relation set in knowledge graph as $\mathcal{E}$ and $\mathcal{R}$ respectively. Each edge in the graph is represented by a triplet $(h, r, t)$, connecting the head entity $h\in\mathcal{E}$ and the tail entity $t\in\mathcal{E}$ with relation $r\in\mathcal{R}$. 
In KG embedding models, entities and relations are mapped to vectors: $\mathcal{E}\rightarrow \mathbb{R}^{d_\mathcal{E}}, \mathcal{R}\rightarrow \mathbb{R}^{d_\mathcal{R}}$.
Here $d_\mathcal{E}, d_\mathcal{R}$ refer to the dimensionality of entity and relation embeddings, respectively.
Specifically, the mapping is learnt via optimizing a defined scoring function $\mathbb{R}^{d_\mathcal{E}}\times\mathbb{R}^{d_\mathcal{R}}\times\mathbb{R}^{d_\mathcal{E}}\rightarrow\mathbb{R}$ measuring the likelihood of triplets \cite{chami2020low}, while maximizing such likelihood only for true triplets.

\xhdr{Hierarchies in knowledge graphs}
Many real-world knowledge graphs contain hierarchical relations~\cite{nickel2017poincare,vilnis2018probabilistic,chami2019hyperbolic}. Such hierarchical structure is characterized by very few top-level nodes corresponding to general and abstract concepts and a vast number of bottom-level nodes corresponding to concrete instances or components of the concept. Examples of hierarchical relations include \textit{isA}, \textit{partOf}.
Note that there may exist multiple (heterogeneous) hierarchical relations in the same graph, which induce several different potentially incompatible hierarchies (i.e., partial orderings) over the same set of entities (Figure~\ref{fig:hete_hier}). In contrast to prior work, our approach is able to model many simultaneous hierarchies over the same set of entities. 

\xhdr{Hyperbolic embeddings}
Hyperbolic embeddings can naturally capture hierarchical structures.
Hyperbolic geometry is a non-Euclidean geometry with a constant negative curvature, where
curvature measures how a geometric manifold deviates from Euclidean space. 
In this work, we use Poincar\'e ball model with constant curvature $c=-1$ as the hyperbolic space for entity embeddings \cite{nickel2017poincare}.
We also investigate on more flexible curvatures, see Appendix~\ref{sec:strategy}, results show that our model is robust enough with constant curvature $c=-1$.
In particular, we denote $d$-dimensional Poincar\'e ball centered at origin as $\mathcal{B}^d = \{\mathbf{x}\in\mathbb{R}^d:\left\|\mathbf{x}\right\|<1\}$, where $\left\|\cdot\right\|$ is the Euclidean norm. The Poincar\'e ball model of hyperbolic space is equipped with Riemannian metric:
\begin{equation}
g^\mathcal{B} = (\frac{2}{1-\left\|\mathbf{x}\right\|^2})^2 g^{E}
\end{equation}
where $g^{E}$ denotes the Euclidean metric, i.e., $g^E=\mathbf{I}_d$. 
The mobius addition $\oplus$ \cite{ganea2018hnn} defined on Poincar\'e ball model with $-1$ curvature is given by:
\begin{equation}
\mathbf{x} \oplus \mathbf{y} = \frac{(1+2\langle \mathbf{x}, \mathbf{y}\rangle+\left\|\mathbf{y}\right\|^2)\mathbf{x}+(1-\left\|\mathbf{x}\right\|^2)\mathbf{y}}{1+2\langle \mathbf{x}, \mathbf{y}\rangle+\left\|\mathbf{x}\right\|^2\left\|\mathbf{y}\right\|^2}
\end{equation}
For each point $\mathbf{x}\in \mathcal{B}^d$, the tangent space $\mathcal{T}_\mathbf{x}\mathcal{B}$ is the Euclidean vector space containing all tangent vectors at $\mathbf{x}$. One can map vectors in $\mathcal{T}_\mathbf{x}\mathcal{B}$ to vectors in $\mathcal{B}^d$ through exponential map $\exp_\mathbf{x}(\cdot):\mathcal{T}_\mathbf{x}\mathcal{B}\rightarrow\mathcal{B}^d$ as follows:
\begin{equation}
\exp_\mathbf{x}(\mathbf{u}) = \mathbf{x} \oplus \tanh(\frac{\left\|\mathbf{u}\right\|}{1-\left\|\mathbf{x}\right\|})\frac{\mathbf{u}}{\left\|\mathbf{u}\right\|}
\end{equation}
Conversely, the logarithmic map $\log_\mathbf{x}(\cdot):\mathcal{B}^d\rightarrow\mathcal{T}_\mathbf{x}\mathcal{B}$ maps vectors in $\mathcal{B}^d$ back to vectors in $\mathcal{T}_\mathbf{x}\mathcal{B}$, in particular:
\begin{equation}
\log_\mathbf{x}(\mathbf{u}) = (1-\left\|\mathbf{x}\right\|)\cdot\tanh^{-1}(\left\|-\mathbf{x}\oplus \mathbf{v}\right\|)\frac{-\mathbf{x}\oplus \mathbf{v}}{\left\|-\mathbf{x}\oplus \mathbf{v}\right\|}
\end{equation}
Also, the hyperbolic distance between $\mathbf{x}, \mathbf{y}\in \mathcal{B}^d$ is:
\begin{equation}
d_\mathcal{B}(\mathbf{x}, \mathbf{y}) = 2\tanh^{-1}(\left\|-\mathbf{x}\oplus\mathbf{y}\right\|)
\end{equation}

A key property of hyperbolic space is that the amount of space covered by a ball of radius $r$ in hyperbolic space increases exponentially with respect to $r$, rather than polynomially as in Euclidean space. This property contributes to the fact that hyperbolic space can naturally model hierarchical tree-like structure.

\xhdr{Hyperbolic entailment cones}
Each hierarchical relation induces a partial ordering over the entities. To capture a given partial ordering, we use the hyperbolic entailment cones \cite{ganea2018hyperbolic}. Figure~\ref{fig:cone} gives an example of 2D hyperbolic cones.

Let $\mathcal{C}_\mathbf{x}$ denotes the cone at apex $\mathbf{x}$. The goal is to model partial order by containment relationship between cones, in particular, the entailment cones satisfy transitivity:
\begin{equation}
\forall \mathbf{x}, \mathbf{y}\in \mathcal{B}^d\backslash \{\mathbf{0}\}:\ \mathbf{y}\in \mathcal{C}_\mathbf{x}\Rightarrow \mathcal{C}_\mathbf{y}\subseteq\mathcal{C}_\mathbf{x}
\label{eq:transitivity}
\end{equation}
Also, for $\mathbf{x}, \mathbf{y}\in\mathcal{B}^d$, we define the angle of $\mathbf{y}$ at $\mathbf{x}$ to be the angle between the half-lines $\overrightarrow{\mathbf{o}\mathbf{x}}$ and $\overrightarrow{\mathbf{x}\mathbf{y}}$ and denote it as $\angle_\mathbf{x}\mathbf{y}$. It can be expressed as:
\begin{equation}
\angle_\mathbf{x}\mathbf{y} = \cos^{-1}(\frac{\langle\mathbf{x}, \mathbf{y}\rangle(1+\left\|\mathbf{x}\right\|^2)-\left\|\mathbf{x}\right\|^2(1+\left\|\mathbf{y}\right\|^2)}{\left\|\mathbf{x}\right\|\left\|\mathbf{x}-\mathbf{y}\right\|\sqrt{1+\left\|\mathbf{x}\right\|^2\left\|\mathbf{y}\right\|^2-2\langle\mathbf{x}, \mathbf{y}\rangle}})
\end{equation}
To satisfy transitivity of nested angular cones and symmetric conditions \cite{ganea2018hyperbolic}, we have the following expression of Poincar\'e entailment cone at apex $\mathbf{x}\in\mathcal{B}^d$:
\begin{equation}
\mathcal{C_{\mathbf{x}}} = \{\mathbf{y}\in\mathcal{B}^d\arrowvert\angle_\mathbf{x}\mathbf{y}\leq \sin^{-1}(K\frac{1-\left\|\mathbf{x}\right\|^2}{\left\|\mathbf{x}\right\|})\}
\label{eq:cone}
\end{equation}
where $K\in\mathbb{R}$ is a hyperparameter (we take $K=0.1$). 
This implies that the half aperture $\phi_\mathbf{x}$ of cone $\mathcal{C_{\mathbf{x}}}$ is as follows:
\begin{equation}
\phi_\mathbf{x} = \sin^{-1}(K\frac{1-\left\|\mathbf{x}\right\|^2}{\left\|\mathbf{x}\right\|})
\end{equation}

\begin{figure}[t]
\centering
\vspace{-18mm}
\includegraphics[scale=0.45]{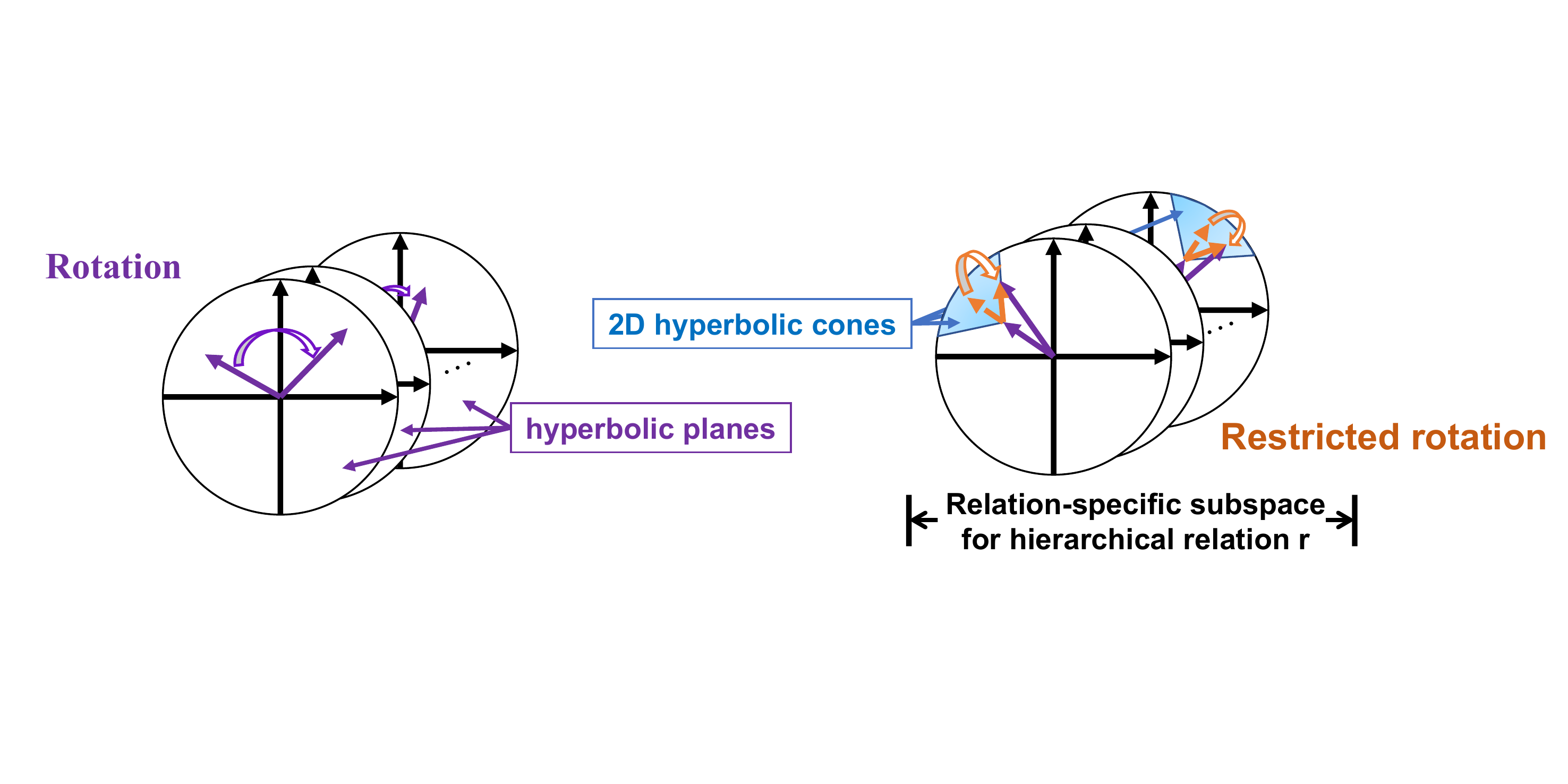}
\vspace{-18mm}
\caption{\name model overview: Embedding space is the product space of $d$ hyperbolic planes and \name learns a different transformation in each hyperbolic plane. \name uses \textbf{restricted rotation} in an assigned relation-specific subspace to model each hierarchical relation $r$ and enforces cone containment constraint in the subspace so that partial ordering of cones is preserved in the subspace. 
For hyperbolic planes not in the subspace, we use a general \textbf{rotation} to model $r$. How to choose a relation-specific subspace for each hierarchical relation is essential and further explained in Sec.~\ref{sec:hete_hierarchies}.
}
\label{fig:cone_overview}
\end{figure}

\subsection{\name Embedding Space and Transformations}
We first introduce the embedding space that \name operates in, and the transformations used to model hierarchical as well as non-hierarchical relations. 

For ease of discussion let's assume that the relation type is given {\em a priori}. In fact, knowledge about hierarchical relations (i.e., transitive closure) is explicitly available in the definition of the relation in KGs such as ConceptNet~\cite{speer2017conceptnet}, WordNet~\cite{miller1995wordnet} and Gene Ontology~\cite{michael2000gene}. When such information is not available, \name can infer ``hierarchicalness'' of a relation by a simple criteria with slight modification to the Krackhardt scores~\cite{krackhardt1994graph}, see Appendix~\ref{krackhardt}. 

\xhdr{Embedding space} The embedding space of \name, $\mathcal{S}$, is a product space of $d$ hyperbolic planes \cite{gu2018learning}, resulting in a total embedding dimension of $2d$.
$\mathcal{S}$ can be denoted as $\mathcal{S} = \mathcal{B}^2\times\mathcal{B}^2\times\cdots\times\mathcal{B}^2$.
Note that this space is different from RotH embedding space \cite{chami2020low}, which is a single $2d$-dimensional hyperbolic space.
\name's embedding space is critical in modeling ancestor-descendant relationships for heterogeneous KGs, since it is more natural when allocating its subspaces (product space of multiple hyperbolic planes) to heterogeneous hierarchical relations.

We denote the embedding of entity $h\in\mathcal{E}$ as $\mathbf{h} = (\mathbf{h}_1, \mathbf{h}_2, \cdots, \mathbf{h}_d)$ where $\mathbf{h}_i\in\mathcal{B}^2$ is the apex of the $i$-th 2D hyperbolic cone.
We model relation $r$ as a cone transformation on each hyperbolic plane from head entity cone to tail entity cone.
Let $\mathbf{r}=(\mathbf{r}_1, \mathbf{r}_2, \cdots, \mathbf{r}_d)$ be the representation of relation $r$. We use $\mathbf{r}_i = (s_i, \theta_i)$ to parameterize transformation for the $i$-th hyperbolic plane as shown in Figure~\ref{fig:cone_overview}.
$s_i>0$ is the scaling factor indicating how far to go in radial direction and $(\theta_i\cdot\phi_{\mathbf{h}_i}/\pi)$ is the rotation angle restricted by half aperture $\phi_{\mathbf{h}_i}$ ($\theta_i\in[-\pi, \pi)$).
To perform hierarchical tasks such as ancestor-descendant prediction, \name uses nested cones in each hyperbolic plane to model the partial ordering property of hierarchical relations, by the cone containment constraint in Def.~\ref{def:cone_containment}.

\begin{definition}{\textbf{Cone containment constraint.}}
If entity $h$ is an ancestor of $t$, then the cone embedding of $t$ has to reside in that of the entity $h$, i.e., $\mathcal{C}_{\mathbf{t}_i}\subseteq\mathcal{C}_{\mathbf{h}_i}$, $\forall i \in \{1, ... d\}$.
\label{def:cone_containment}
\end{definition}
The cone containment constraint can be enforced in any of the hyperbolic plane components in $\mathcal{S}$.
Next we introduce \name's transformations for characterizing hierarchical and non-hierarchical patterns of relation $r$ in triple $(h, r, t)$. 
Note that we utilize both transformations to model hierarchical relations $r$ to capture non-hierarchical properties, i.e., symmetry, composition, etc, as well as hierarchical properties, i.e., partial ordering.
We do this by performing different transformations in different subspaces of $\mathcal{S}$, as discussed in detail in Sec.~\ref{sec:hete_hierarchies}.

\begin{figure}[t]
\centering
\subfigure[Cone rotation]{
\begin{minipage}{0.4\linewidth}
\centering
\includegraphics[width=1.6in]{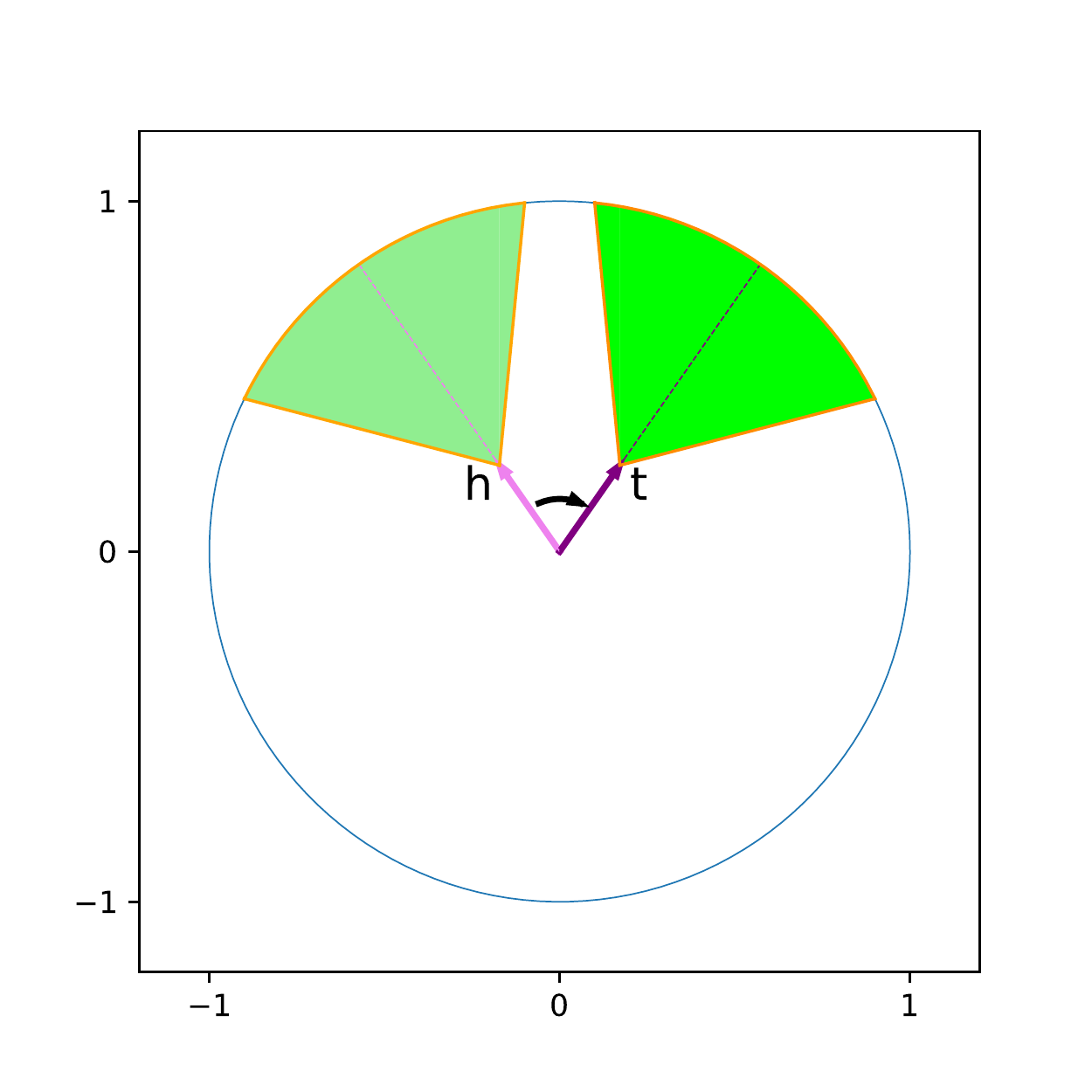}
\label{fig:rotation}
\end{minipage}%
}%
\subfigure[Restricted cone rotation]{
\begin{minipage}{0.4\linewidth}
\centering
\includegraphics[width=1.6in]{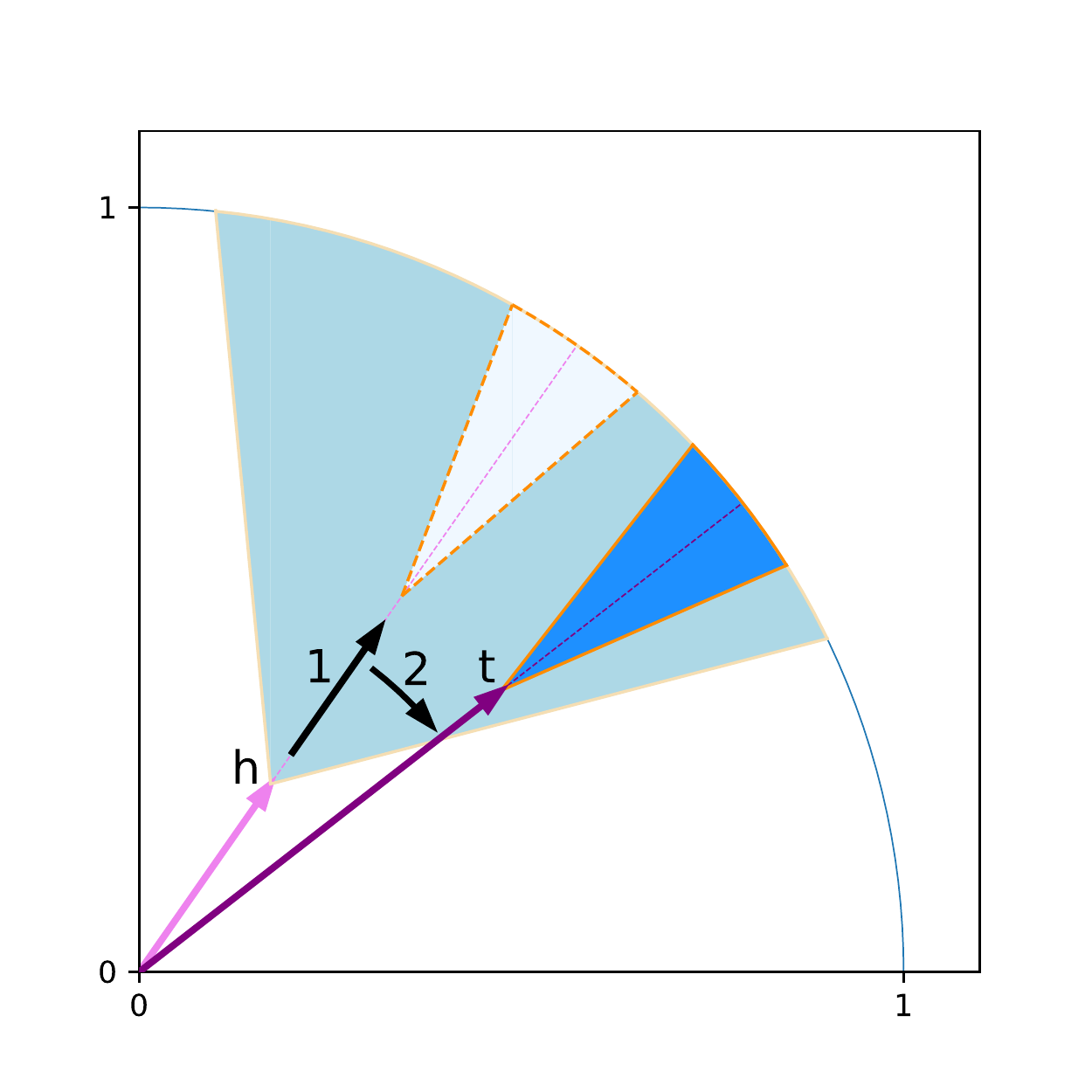}
\label{fig:restricted}
\end{minipage}%
}%
\centering
\caption{Transformations in \name in Poincar\'e ball: (a) Cone rotation from $h$ to $t$ used for non-hierarchical relations; (b) Restricted rotation from the cone of parent $h$ to the cone of child $t$ used for hierarchical relations, where ``1'' corresponds to scaling and ``2'' to rotation ($s_i, \theta_i$) in Eq.~\ref{eq:restricted}.
}
\label{fig:rotation_transformations}
\vspace{-2mm}
\end{figure}

\xhdr{Transformation for modeling non-hierarchical properties}
Rotation is an expressive transformation to capture relation between entities~\cite{sun2019rotate}.
Analogous to RotatE, we adopt \textbf{rotation transformation} $f_1$ to model non-hierarchical properties~(Figure~\ref{fig:rotation}). For rotation in the $i$-th hyperbolic plane, 
\begin{equation}
    f_1(\mathbf{h}_i, \mathbf{r}_i) = \exp_\mathbf{o}(\mathbf{G}(\theta_i)\log_\mathbf{o}(\mathbf{h}_i))
\label{eq:rotation}
\end{equation}
where $\mathbf{G}(\theta_i)$ is the Givens rotation matrix:
\begin{equation}
    \mathbf{G}(\theta_i) = 
    \left[                
  \begin{array}{cc}   
    \cos(\theta_i) & -\sin(\theta_i) \\  
    \sin(\theta_i) & \cos(\theta_i) \\  
  \end{array}
\right]
\end{equation}
We also show that the rotation transformation in Eq.~\ref{eq:rotation} is expressive: It can model relation patterns including symmetry, anti-symmetry, inversion, and composition (Appendix~\ref{sec:proof}).

\xhdr{Transformation for modeling hierarchical properties}
However, $f_1$ cannot be directly applied to model hierarchical relations, because rotation does not obey transitive property: rotation by $\theta_i$ twice will result in a rotation of $2\theta_i$, instead of $\theta_i$. Hence it cannot guarantee $(h_1,r,h_3)$ when $(h_1,r,h_2)$ and $(h_2,r,h_3)$ are true.
We use \textbf{restricted rotation transformation} $f_2$ to model hierarchical relations. We impose cone containment constraint to preserve partial ordering of cones after the transformation.
Without loss of generality we assume relation $r$ is a hyponym type relation, the restricted rotation from $h$ to $t$ in $i$-th hyperbolic plane is as follows (we perform restricted rotation from $t$ to $h$ if $r$ is a hypernym relation):
\begin{equation}
    f_2(\mathbf{h}_i, \mathbf{r}_i) =  \exp_{\mathbf{h}_i}(s_i\cdot\mathbf{G}(\theta_i\frac{\phi_\mathbf{\mathbf{h}_i}}{\pi})\overline{\mathbf{h}}_i), \, \mathbf{r}_i = (s_i, \theta_i)
\label{eq:restricted}
\end{equation}
where $\phi_\mathbf{\mathbf{h}_i}$ is the half aperture of cone $\mathbf{h}_i$. $\overline{\mathbf{h}}_i$ is the unit vector of $\mathbf{h}_i$ in the tangent space of $\mathbf{h}_i$:
\begin{equation}
    \overline{\mathbf{h}}_i = \widehat{\mathbf{h}}_i / ||\widehat{\mathbf{h}}_i||,\ \widehat{\mathbf{h}}_i = \log_{\mathbf{h}_i}(\frac{1+||\mathbf{h}_i||}{2||\mathbf{h}_i||}\mathbf{h}_i)
\end{equation}
Figure~\ref{fig:restricted} illustrates the two-step transformation described in Eq.~\ref{eq:restricted}, namely the scaling step and the rotation step.

\subsection{\name Model of Heterogeneous Hierarchies}
\label{sec:hete_hierarchies}
In the previous section, we explained how we enforce cone containment constraint for hierarchical relations, however two challenges remain when simultaneously modeling multiple heterogeneous hierarchies:
{\bf (1) Partial ordering:}
Suppose that there is a hyponym relation between entities $h_1$ and $h_2$, and a \emph{different} hyponym relation between entities $h_2$ and $h_3$. Then a na\"ive model would enforce that the cone of $h_1$ contains the cone of $h_2$ which contains the cone of $h_3$, implying that a hyponym relation exists between $h_1$ and $h_3$, which is not correct. {\bf (2) Expressive power:} Cone containment constraint, while ensuring hierarchical structure by geometric entailment, limits the set of possible rotation transformations and thus limits the model's expressive power. 

To address these challenges we proceed as follows. Instead of enforcing cone containment constraint in the entire embedding space, \name proposes a novel technique to assign unique subspace for each hierarchical relation, \textit{i.e.} we enforce cone containment constraint only in a subset of $d$ hyperbolic planes. Next we further elaborate on this idea.

In particular, for a hierarchical relation $r$, we assign a corresponding subspace of $\mathcal{S}$, which is a product space of a subset of hyperbolic planes. 
Then, we use restricted rotation in the subspace and rotation in the complement space.
We train \name to enforce cone containment constraint in the relation-specific subspace. 
The subspace can be represented by a $d$-dimensional mask $\mathbf{m}, \mathbf{m}_i\in\{0, 1\}$, and $\mathbf{m}_i=1$ indicates that cone containment is enforced in the $i$-th hyperbolic plane. 
We then extend such notation to all relations where $\mathbf{m}=\mathbf{0}$ for non-hierarchical relations. 

Our design of leveraging both transformations to model hierarchical relations is crucial in that they capture different aspects of the relation. The use of restricted rotation along with cone containment constraint serves to preserve partial ordering of a hierarchical relation in its relation-specific subspace. But restricted rotation alone is insufficient: hierarchical relations also possess other properties such as composition and symmetry that cannot be modeled by restricted rotation. Hence we augment with the rotation transformation to capture these properties, allowing composition of different hierarchical and non-hierarchical relations through rotations in the complement space.
We further provide theoretical and empirical results in Appendix~\ref{sec:design} to support that both transformations are of great significance to the expressiveness of our model.

Putting it all together gives us the following distance scoring function (we use $(v_i)_{i\in\{1,\cdots,d\}}$ in the following to denote a $d$-dimensional vector $\mathbf{v}$):
\begin{equation}
\begin{aligned}
\label{eq:score}
    \psi(h, r, t) = -\frac{1}{d}[ & \mathbf{m}\cdot(d_\mathcal{B}(f_2(\mathbf{h}_i, \mathbf{r}_i), \mathbf{t}_i))_{i\in\{1, \cdots, d\}}  \\ 
    +&(\mathbf{1}-\mathbf{m})\cdot(d_\mathcal{B}(f_1(\mathbf{h}_i, \mathbf{r}_i), \mathbf{t}_i))_{i\in\{1, \cdots, d\}}] + b_h + b_t
\end{aligned}
\end{equation}
where the first term corresponds to the restricted rotation in relation-specific subspace, and the second term corresponds to the rotation in complementary space.  A high score indicates that cone of entity $h$ after relation-specific transformation $r$ is close to the cone of entity $t$ in terms of hyperbolic distance $d_\mathcal{B}$. 
Note that $b_h, b_t$ are the learnt radius parameters of $h, t$ which can be interpreted as margins \cite{balazevic2019multi}.

\xhdr{Subspace allocation} We assign equal dimensional subspaces for all hierarchical relations. We discuss and compare several strategies in assigning subspaces for hierarchical relations in Appendix~\ref{sec:strategy}, including whether to use overlapping subspaces or orthogonal subspaces for different hierarchical relations, as well as the choice of dimensionality of subspaces.
Overlapping subspaces (Appendix~\ref{sec:strategy}) allow the model to perform well and enable it to scale to knowledge graphs with a large number of relations, since there are exponentially many possible overlapping subspaces that can potentially correspond to different hierarchical relations.

\subsection{\name Loss Function}
We use a loss function composed of two parts. The first part of the loss function aims to ensure that for a given head entity $h$ and relation $r$ the distance to the true tail entity $t$ is smaller than to the negative tail entity $t'$:
\begin{equation}
\begin{aligned}
    \mathcal{L}_d(h, r, t) = &-\log\sigma(\psi(h, r, t)) 
    -\sum_{t'\in\mathcal{T}}\frac{1}{|\mathcal{T}|}\log\sigma(-\psi(h, r, t'))
\end{aligned}
\end{equation}
where $(h, r, t)$ denotes a positive training example/triplet, and we generate negative samples $(h,r,t')$ by substituting the tail with a random entity in $\mathcal{T} \subseteq \mathcal{E}$, a random set of entities in KG excluding $t$. 

However, the distance loss $\mathcal{L}_d$ does not guarantee embeddings satisfying the cone containment constraint, since the distance between transformed head embedding and tail embedding can still be non-zero after training. Hence we additionally introduce the \textbf{angle loss} (without loss of generality let $r$ be a hyponym relation):
\begin{equation}
    \mathcal{L}_a(h, r, t) = \mathbf{m} \cdot (\max(0, \angle_{\mathbf{h}_i}\mathbf{t}_i - \phi(\mathbf{h}_i)))_{i\in\{1, \cdots, d\}}
\label{eq:angle}
\end{equation}
which directly encourages cone of $h$ to contain cone of $t$ in relation-specific subspaces, by constraining the angle between the cones.
The final loss is then a weighted sum of the distance loss and the angle loss, where weight $w$ is a hyperparameter (We investigate the choice of $w$ in Appendix~\ref{sec:strategy}):
\begin{equation}
    \mathcal{L} = \mathcal{L}_d + w\cdot\mathcal{L}_a
\label{eq:total}
\vspace{-3mm}
\end{equation}

\begin{table}[t]
\centering
\resizebox{\textwidth}{!}{
\begin{tabular}{c|cccccc}
\hline
Dataset & \#entities & \#relations & \#training & \#validation & \#test & Examples of hierarchical relations \\
\hline
WN18RR & 40,943 & 11 & 86,385 & 3,034 & 3,134 & \emph{hypernym, has part} \\
DDB14 & 9,203 & 14 & 38,233 & 4,000 & 4,000 & \emph{subtype of, subset of} \\
GO21 & 89,127 & 21 & 796,136 & 5,000 & 5,000 & \emph{part of, is a} \\
FB15k-237 & 14,541 & 237 & 272,115 & 17,535 & 20,466 & \emph{location/contains, /music/genre/parent\c genre} \\
\hline
\end{tabular}
}
\vspace{1mm}
\caption{Datasets statistics. Note that FB15k-237 has very few such hierarchical relations.}
\label{tb:statistics}
\vspace{-2mm}
\end{table}
\section{Experiments}
\label{sec:experiments}
Given a KG containing many hierarchical and non-hierarchical relations, our experiments evaluate:
\textbf{(A)} Performance of \name on hierarchical reasoning task of predicting if entity $h_1$ is an ancestor of entity $h_2$.
\textbf{(B)} Performance of \name on generic KG completion tasks.

\xhdr{Datasets}
We use four knowledge graph benchmarks (Table~\ref{tb:statistics}): WordNet lexical knowledge graph (WN18RR \cite{bordes2013translating, dettmers2018convolutional}), drug knowledge graph (DDB14 \cite{wang2020entity}), and a KG capturing common knowledge (FB15k-237 \cite{toutanova2015observed}).
Furthermore, we also curated a new biomedical knowledge graph GO21, which models genes and the hierarchy of biological processes they participate in.

\xhdr{Model training}
During training, we use Adam \cite{kingma2014adam} as the optimizer and search hyperparameters including batch size, embedding dimension, learning rate, angle loss weight and dimension of subspace for each hierarchical relation. (Training details and standard deviations in Appendix~\ref{sec:training}).\footnote{The code of our paper is available at \url{http://snap.stanford.edu/cone}.}

We use a \textbf{single} trained model (without fine-tuning) for all evaluation tasks:
On ancestor-descendant relationship prediction, our scoring function for a pair $(h, t)$ with hierarchical relation $r$ is the angle loss in Eq.~\ref{eq:angle} where a lower score means $h$ is more likely to be an ancestor of $t$. For KG completion task we use the scoring function $\psi(h, r, t)$ in Eq.~\ref{eq:score} to rank the triples.

\subsection{Hierarchical Reasoning: Ancestor-descendant Prediction}
\begin{table}[t]
\centering
\resizebox{\textwidth}{!}{
\begin{tabular}{c|ccc|ccc|ccc}
\hline 
 & \multicolumn{3}{c|}{WN18RR} & \multicolumn{3}{c|}{DDB14} & \multicolumn{3}{c}{GO21} \\
\hline
 & \multicolumn{9}{c}{Fraction of inferred descendant pairs among all true descendant pairs in the test set} \\
\hline
Model & 0\% & 50\% & 100\% & 0\% & 50\% & 100\% & 0\% & 50\% & 100\% \\
\hline
Order \cite{vendrov2015order} & \underline{.889} & \underline{.739} & .498 & .731 & .633 & .513 & \underline{.642} & \underline{.592} & .534 \\
Poincar\'e \cite{nickel2017poincare}  & .810 & .685 & .508 & \underline{.976} & \underline{.832} & .571 & .525 & .519 & .516 \\
HypCone \cite{ganea2018hyperbolic} & .799 & .677 & .504 & .973 & .823 & \underline{.594} & .554 & .539 & .519 \\
RotatE \cite{sun2019rotate} & .601 & .593 & .582  & .615 & .590 & .565 & .546 & .534 & .526 \\
RotH \cite{chami2020low} & .601 & .608 & \underline{.611} & .609 & .596 & .578 & .596 & .583 & \underline{.564} \\
\hline
ConE & {\bf .895} & {\bf .801} & {\bf .679} & {\bf .981} & {\bf .909} & {\bf .818} & {\bf .789} & {\bf .744} & {\bf .693} \\
\hline
Improvement (\%) & +1.9\% & +9.6\% & +11.1\% & +0.5\% & +10.3\% & +38.4\% & +22.9\% & +25.7\% & +22.9\% \\
\hline
\end{tabular}
}
\vspace{1mm}
\caption{Ancestor-descendant prediction results in mAP (mean average precision).  Best score in \textbf{bold} and second best \underline{underlined}. We create different test sets that get harder as they contain more and more test cases (0\%, 50\%, 100\%) of inferred descendant pairs.
}
\label{tb:map}
\vspace{-2mm}
\end{table}

Next we define ancestor-descendant relationship prediction task to test model's ability on hierarchical reasoning. 
Given two entities, the goal makes a binary prediction if they have ancestor-descendant relationship: 
\begin{definition}{\textbf{Ancestor-descendant relationship.}}
Entity pair $(h_1, h_2)$ is considered to have ancestor-descendant relationship if: there exists a path from $h_1$ to $h_2$ that only contains one type of hyponym relation, or a path from $h_2$ to $h_1$ that only contains one type of hypernym relation.
\label{def:ancestor_descendant}
\end{definition}
Our evaluation setting is a generalization of the transitive closure prediction \cite{vendrov2015order,nickel2017poincare,ganea2018hyperbolic} which is defined only over a single hierarchy, but our knowledge graphs contain multiple hierarchies (hierarchical relations). 
More precisely: \textbf{(1)} When heterogeneous hierarchies coexist in the graph, we compute the transitive closure induced by each hierarchical relation separately. The test set for each hierarchical relation is a random collection sampled from all transitive closures of that relation.
\textbf{(2)} To increase the difficulty of the prediction task, our evaluation also considers \textbf{inferred descendant pairs}, which are only possible to be inferred when simultaneously considering hierarchical and non-hierarchical relations in KG, due to missing links in KG. We call a descendant pair $(u, v)$ an inferred descendant pair if their ancestor-descendant relationship can be inferred from the whole graph but {\em not} from the training set. For instance, \textit{(Tree,WinePalm)} would be an {\em inferred descendant pair} if the \textit{subClass} relation between \textit{Tree} and \textit{PalmTree} is missing in training set. 
We construct the inferred descendant pairs by taking the transitive closures of the entire graph, and exclude the transitive closures of relations in the training set.
In our experiments, we consider three test settings: 0\%, 50\%, 100\%, corresponding to the fraction of inferred descendant pairs among all true descendant pairs in the test set, and the setting with a higher fraction is harder.

On each dataset, we extract 50k ancestor-descendant pairs. 
For each pair, we randomly replace the true descendant with a random entity in the graph, resulting in a total of 100k pairs.
Our way of selecting negative examples offsets the bias during learning that is prevalent in baseline models: the models tend to always give higher scores to pairs with a high-level node as ancestor, since high-level nodes usually have more descendants presented in training data. We replace the true descendant while keeping the true ancestor unchanged for the negative sample, and thus the model will not be able to ``cheat'' by taking advantage of the fore-mentioned bias.
For each model, we then use its scoring function to rank all the pairs.
We use the standard mean average precision (mAP) to evaluate the performance on this binary classification task. 
We further show the AUROC results in Appendix~\ref{sec:AUC}.

\xhdr{Baselines}
We compare our method with state-of-the-art methods for hierarchical reasoning, including Order embeddings \cite{vendrov2015order}, Poincar\'e embeddings \cite{nickel2017poincare} and Hyperbolic entailment cones \cite{ganea2018hyperbolic}. Note that these methods can only handle a single hierarchical relation at a time. So each baseline trains a separate embedding for each hierarchical relation and then learns a scoring function on the embedding of the two entities. To ensure that the experiment controls the model size, we enforce that in baselines, the sum of embedding dimensions of all relations is equal to the relation embedding dimension of \name.
We also perform comprehensive hyperparameter search for all baselines (Appendix~\ref{sec:training}).
Although KG embedding models (RotatE~\cite{sun2019rotate} and RotH~\cite{chami2020low}) cannot be directly applied to this task, we adapt them to perform this task by separately training an MLP to make binary classification on ancestor-descendant pair, taking the concatenation of the two entity embeddings as input. Note that \name outperforms these KG completion methods without even requiring additional training.

\begin{figure}[t]
\centering
\subfigure[Visualization of RotH's embedding]{
\begin{minipage}{0.45\linewidth}
\centering
\includegraphics[height=1.6in]{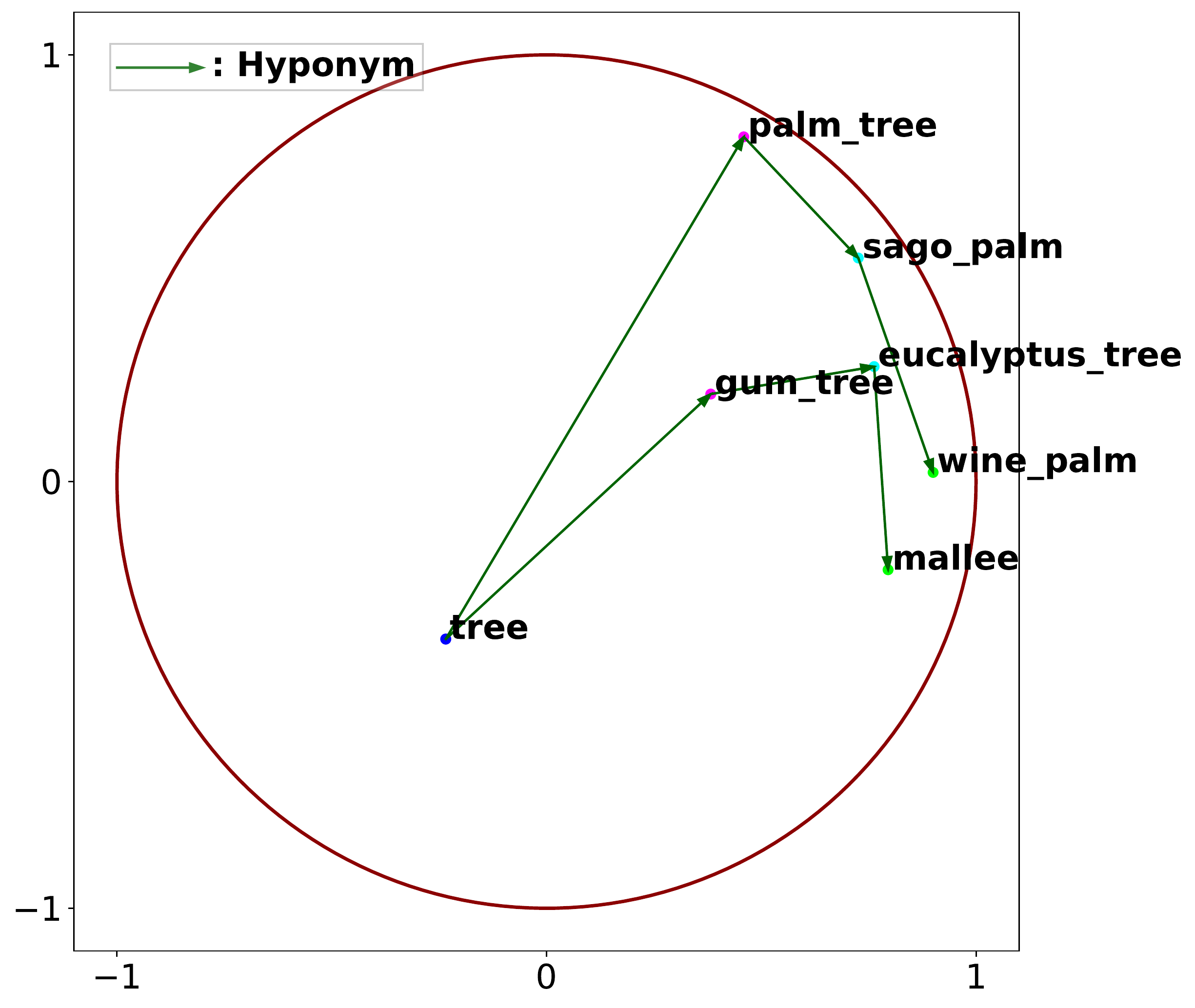}
\label{fig:vis-RotH}
\end{minipage}%
}%
\qquad
\subfigure[Visualization of ConE's embedding]{
\begin{minipage}{0.45\linewidth}
\centering
\includegraphics[height=1.6in]{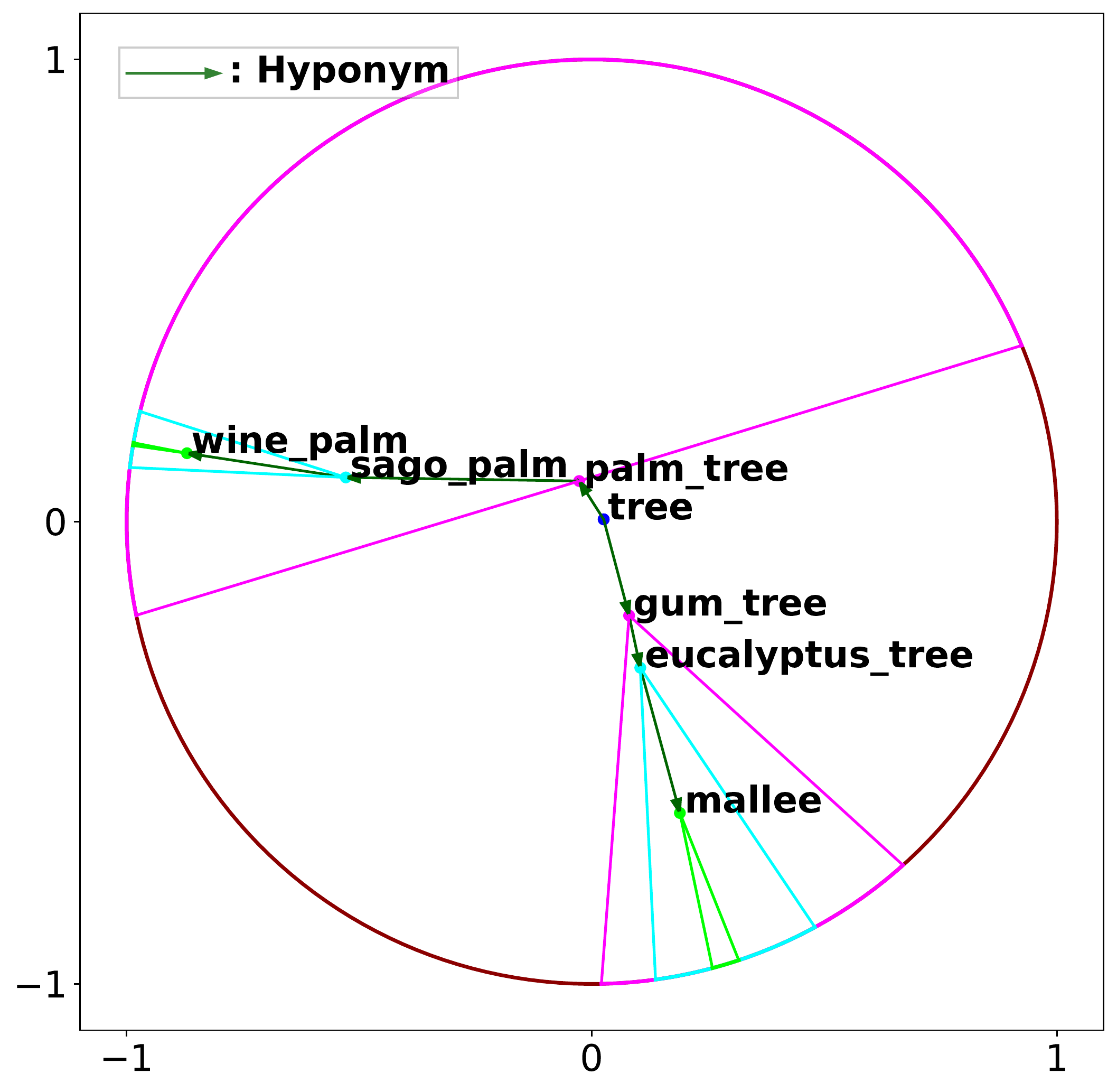}
\label{fig:vis-ConE}
\end{minipage}%
}%
\centering
\caption{The embeddings of RotH and \name, trained on WN18RR, projecting to one hyperbolic plane. We show the embedding of a family of trees, and the arrows point from higher level entities to lower level entities, representing the hierarchical relation ``\textit{Hyponym}''. Different levels of entities and their corresponding cones in \name model (Figure~\ref{fig:vis-ConE}) are marked with different colors. In \name model, the embeddings of high-level entities (e.g., tree, palm tree) are close to the center of the hyperbolic plane, while embeddings of their descendant entities (e.g., wine palm, mallee) fall in their hyperbolic cones.}
\label{fig:vis}
\vspace{-4mm}
\end{figure}

\xhdr{Results}
Table~\ref{tb:map} reports the ancestor-descendant prediction results of \name and the baselines. 
We observe that the novel subspace transformation of \name results in its superior performance in this task. 
Our model consistently outperforms baseline methods on all three datasets.
As we expected, KG embedding models cannot perform well on this task (in the range of $0.5\sim0.6$ across all settings), since they do not explicitly model the partial ordering property of the hierarchical relations.
In contrast, our visualization of \name's embedding in Figure~\ref{fig:vis} suggests that \name faithfully preserves the cone containment constraint in modeling hierarchical relations, while RotH's embedding exhibit less hierarchical structure.
As a result, \name simultaneously captures the heterogeneous relation modeling and partial ordering, combining the best of both worlds.
Our improvement is more significant as the fraction of inferred descendant pairs increases. 
This shows that \name not only embeds a given hierarchical structure, but can also infer missing hierarchical links by modeling other non-hierarchical relations at the same time.
Thanks to the restricted rotation transformation and the use of product spaces of hyperbolic planes, \name can faithfully model the hierarchies without requiring all transitive closures in the training set. 
We further perform additional studies to explore reasons for the performance of each method on ancestor-descendant prediction task in Appendix~\ref{sec:AUC}.

\xhdr{Lowest common ancestor prediction task} Moreover, we demonstrate flexibility and power of \name using a hierarchical analysis task: lowest common ancestor (LCA) prediction, which requires both the ability to model ancestor-descendant relationship and to distinguish the lowest ancestor. Results show that \name can precisely predict LCA, outperforming over 100\% on Hits@3 and Hits@10 metrics compared to previous methods (See detailed results and analysis in Appendix~\ref{sec:LCA}).

\subsection{Knowledge Graph Completion}

We also experiment on knowledge graph completion task where missing links include hierarchical relations as well as non-hierarchical relations. We follow the standard evaluation setting~\cite{bordes2013translating}.

\xhdr{Baselines}
We compare \name model to state-of-the-art models on knowledge graph completion task, including TransE \cite{bordes2013translating}, RotatE \cite{sun2019rotate}, TuckER \cite{balavzevic2019tucker} and HAKE \cite{zhang2020learning}, as well as MuRP \cite{balazevic2019multi} and RotH \cite{chami2020low}, which both operate on a hyperbolic space. 

\xhdr{Results}
Table~\ref{tb:link} reports the KG completion results.
Over the first three hierarchical datasets considered, \name achieves state-of-the-art results over many recent baselines, including the recently proposed hyperbolic approaches RotH and MuRP.
We also notice that the margins on Hits@1 and Hits@3 scores are much larger than Hits@10, indicating that our model provides the most accurate predictions.
We further use Krackhardt scores $\kappa$ to measure how hierarchical each graph is~\cite{krackhardt1994graph}. The score consists of four metrics (\emph{(connectedness, hierarchy, efficiency, LUBedness)}, Appendix~\ref{krackhardt}), where if a graph is maximally hierarchical (i.e., a tree) then its Krackhardt score is $(1, 1, 1, 1)$, and higher score on four metrics indicate a more hierarchical structure. Notice that the Krackhardt scores of FB15k-237 are approximately three times lower than those of WN18RR, DDB14 and GO21, indicating that FB15k-237 is indeed non-hierarchical.
We can see that our \name model still performs better than other hierarchical KG embedding models (RotH and MuRP) on FB15k-237 and is comparable to SOTA model (TuckER).
Overall, this shows that \name can scale to a large number of relations, and that it has competitive performance even in non-hierarchical knowledge graphs.

We further analyze the performance of \name in low-dimensional regimes in Appendix~\ref{sec:low}. Similar to previous studies, the hyperbolic-space-based \name model performs much better than Euclidean KG embeddings in low dimensions ($d=32$). \name performs similar to previous hyperbolic KG embedding baselines in low dimensions, but outperforms them in high-dimensional regimes (Table~\ref{tb:map}).

\xhdr{Ablation study} 
We further compare the performance of our model with one that does not use cone restricted rotation for modeling hierarchical relations and one that does not use rotation for modeling hierarchical relations. 
Ablation results suggest that both transformations, i.e., cone restricted rotation and rotation, are critical in predicting missing hierarchical relations (Appendix~\ref{sec:ablation}). 
In particular, our ablation results on each individual hierarchical relation suggest that with cone restricted rotation, \name can simultaneously model heterogeneous hierarchical relations effectively.

\begin{table}[t]%
\centering  
\resizebox{\textwidth}{!}{
\begin{tabular}{c|cccc|cccc|cccc|cccc}
\hline
 & \multicolumn{4}{c}{WN18RR} & \multicolumn{4}{|c}{DDB14} & \multicolumn{4}{|c}{GO21} & \multicolumn{4}{|c}{FB15k-237} \\
 & \multicolumn{4}{c}{$\kappa=(1.00, 0.61, 0.99, 0.50)$} & \multicolumn{4}{|c}{$\kappa=(1.00, 0.84, 0.78, 0.18)$} & \multicolumn{4}{|c}{$\kappa=(1.00, 0.65, 0.96, 0.22)$} & \multicolumn{4}{|c}{$\kappa=(1.00, 0.18, 0.36, 0.06)$} \\
\hline
Model & MRR & H@1 & H@3 & H@10 & MRR & H@1 & H@3 & H@10 & MRR & H@1 & H@3 & H@10 & MRR & H@1 & H@3 & H@10 \\
\hline
TransE \cite{bordes2013translating} & .226 & .017 & .403 & .532 & .183 & .103 & .212 & .337 & .149 & .066 & .179 & .310 & .294 & - & - & .465 \\  
RotatE \cite{sun2019rotate} & .476 & .428 & .429 & .571 & \underline{.225} & \underline{.154} & \underline{.245} & \underline{.362} & .203 & .123 & \underline{.234} & {\bf .357} & .338 & .241 & .375 & .533 \\  
TuckER \cite{balavzevic2019tucker} & .470 & .443 & .482 & .526 & .198 & .137 & .219 & .314 & \underline{.205} & \underline{.136} & .222 & .342 & {\bf .358} & {\bf .266} & {\bf .394} & {\bf .544} \\
HAKE \cite{zhang2020learning} & \underline{.496} & \underline{.451} & .513 & \underline{.582} & .217 & .146 & .237 & .361 & .169 & .104 & .185 & .295 & .341 & .243 & .378 & .535 \\
MuRP \cite{balazevic2019multi} & .481 & .440 & .495 & .566 & .214  & .146  & .231  & .349  & .166 & .100  & .181  & .301 & .335 & .243 & .367 & .518 \\
RotH \cite{chami2020low} & .495 & .449 & \underline{.514} & {\bf .586} & .223 & .152 & .245 & .357 & .151 & .079 & .171 & .289 & .344 & .246 & .380 & .535 \\
\hline
ConE & \underline{.496} & {\bf .453} & {\bf .515} & .579 & {\bf .231} & {\bf .161} & {\bf .252} & {\bf .364}  & {\bf .211} & {\bf .140} & {\bf .237} & \underline{.347} & \underline{.345} & \underline{.247} & \underline{.381} & \underline{.540} \\
\hline  
\end{tabular}
}
\vspace{1mm}
\caption{Knowledge graph completion results, best out of dimension $d\in\{100,250,500\}$. Best score in \textbf{bold} and second best \underline{underlined}.
$\kappa$ is a tuple denoting the 4 Krackhardt scores \cite{krackhardt1994graph} that measure how hierarchical a graph is, higher scores mean more hierarchical. \name achieves the best MRR and Hits@1 results in hierarchical KGs.
}
\label{tb:link}
\vspace{-4mm}
\end{table}

\section{Conclusion}
\label{sec:conclusion}
In this paper, we propose \name, a hierarchical KG embedding method that models entities as hyperbolic cones and uses different transformations between cones to simultaneously capture hierarchical and non-hierarchical relation patterns. 
We apply cone containment constraint to relation-specific subspaces to capture hierarchical information in heterogeneous knowledge graphs.
\name can simultaneously perform knowledge graph completion task and hierarchical task, and achieves state-of-the-art results on both tasks across three hierarchical knowledge graph datasets.

\begin{ack}

We gratefully acknowledge the support of
DARPA under Nos. HR00112190039 (TAMI), N660011924033 (MCS);
ARO under Nos. W911NF-16-1-0342 (MURI), W911NF-16-1-0171 (DURIP);
NSF under Nos. OAC-1835598 (CINES), OAC-1934578 (HDR), CCF-1918940 (Expeditions), IIS-2030477 (RAPID),
NIH under No. R56LM013365;
Stanford Data Science Initiative, 
Wu Tsai Neurosciences Institute,
Chan Zuckerberg Biohub,
Amazon, JPMorgan Chase, Docomo, Hitachi, Intel, JD.com, KDDI, NVIDIA, Dell, Toshiba, Visa, and UnitedHealth Group. 
Hongyu Ren is supported by the Masason Foundation Fellowship and the Apple PhD Fellowship. Jure Leskovec is a Chan Zuckerberg Biohub investigator.

The content is solely the responsibility of the authors and does not necessarily represent the official views of the funding entities.

\end{ack}

\bibliography{refs}
\bibliographystyle{ieeetr}

\newpage
\appendix
\onecolumn

\appendix
\section{Theoretical and empirical evidence for \name's design choice}
\label{sec:design}
Here we provide theoretical and empirical results to support that \name's design choice makes sense, i.e., both rotation transformation and restricted transformation play a crucial role to the expressiveness of the model.
\subsection{Proof for transformations}
\label{sec:proof}
\subsubsection{Proof for rotation transformation}
We will show that the rotation transformation in Eq.~\ref{eq:rotation} can model all relation patterns that can be modeled by its Euclidean counterpart RotatE \cite{sun2019rotate}.

Three most common relation patterns are discussed in \cite{sun2019rotate}, including symmetry pattern, inverse pattern and composition pattern. Let $\mathbb{T}$ denote the set of all true triples. We formally define the three relation patterns as follows.
\begin{definition}
If a relation $r$ satisfies symmetric pattern, then 
\begin{displaymath}
\forall h, t\in\mathcal{E},\ (h, r, t)\in\mathbb{T}\Rightarrow(t, r, h)\in\mathbb{T}
\end{displaymath}
\end{definition}
\begin{definition}
If relation $r_1$ and $r_2$ satisfies inverse pattern, i.e., $r_1$ is inverse to $r_2$, we have
\begin{displaymath}
\forall h, t\in\mathcal{E},\ (h, r_1, t)\in\mathbb{T}\Rightarrow(t, r_2, h)\in\mathbb{T}
\end{displaymath}
\end{definition}
\begin{definition}
If relation $r_1$ is composed of $r_2$ and $r_3$, then they satisfies composition pattern, 
\begin{displaymath}
\forall h, m, t\in\mathcal{E},\ (h, r_2, m)\in\mathbb{T}\land(m, r_3, t)\in\mathbb{T}\Rightarrow(h, r_1, t)\in\mathbb{T}
\end{displaymath}
\end{definition}

\begin{theorem}
Rotation transformation can model symmetric pattern.
\label{thm:1}
\end{theorem}
\xproof
If $r$ is a symmetric relation, then for each triple $(h, r, t)$, its symmetric triple $(t, r, h)$ is also true. For $i\in\{1,2,\cdots,d\}$, we have
\begin{displaymath}
    \mathbf{t}_i = \exp_\mathbf{o}(\mathbf{G}(\theta_i)\log_\mathbf{o}(\mathbf{h}_i)),\ \mathbf{h}_i = \exp_\mathbf{o}(\mathbf{G}(\theta_i)\log_\mathbf{o}(\mathbf{t}_i))
\end{displaymath}
Let $\mathbf{I}$ denote the identity matrix. By taking logarithmic map on both sides, we have
\begin{displaymath}
    \log_\mathbf{o}(\mathbf{t}_i) = \mathbf{G}(\theta_i)\log_\mathbf{o}(\mathbf{h}_i),\ \log_\mathbf{o}(\mathbf{h}_i) = \mathbf{G}(\theta_i)\log_\mathbf{o}(\mathbf{t}_i)\ \Rightarrow\ 
    \mathbf{G}^2(\theta_i) = \mathbf{I}
\end{displaymath}
which holds true when $\theta_i=-\pi$ or $\theta_i=0$ (still we assume $\theta_i\in[-\pi,\pi)$).

\begin{theorem}
Rotation transformation can model inverse pattern.
\end{theorem}
\xproof
If $r_1$ and $r_2$ are inverse relations, then for each triple $(h, r_1, t)$, its inverse triple $(t, r_2, h)$ also holds. Let $(\theta_i)_{i\in\{1,\cdots,d\}}$ denote the rotation parameter of relation $r_1$ and $(\alpha_i)_{i\in\{1,\cdots,d\}}$ denote the rotation parameter of relation $r_2$. Similar to the proof of Theorem~\ref{thm:1}, we take logarithmic map on rotation transformation, then
\begin{displaymath}
    \log_\mathbf{o}(\mathbf{t}_i) = \mathbf{G}(\theta_i)\log_\mathbf{o}(\mathbf{h}_i),\ 
    \log_\mathbf{o}(\mathbf{h}_i) = \mathbf{G}(\alpha_i)\log_\mathbf{o}(\mathbf{t}_i)\ \Rightarrow\ 
    \mathbf{G}(\theta_i)\mathbf{G}(\alpha_i) = \mathbf{I}
\end{displaymath}
which holds true when $\theta_i+\alpha_i=0$.

\begin{theorem}
Rotation transformation can model composition pattern.
\end{theorem}
\xproof
If relation $r_1$ is composed of $r_2$ and $r_3$, then triple $(h, r_1, t)$ exists when $(h, r_2, m)$ and $(m, r_3, t)$ exist. Let $(\theta_i)_{i\in\{1,\cdots,d\}}$, $(\alpha_i)_{i\in\{1,\cdots,d\}}$, $(\beta_i)_{i\in\{1,\cdots,d\}}$, denote their rotation parameters correspondingly. Still we take logarithmic map on rotation transformation and it can be derived that
\begin{displaymath}
\begin{aligned}
    \log_\mathbf{o}(\mathbf{t}_i) = \mathbf{G}(\theta_i)\log_\mathbf{o}(\mathbf{h}_i),\ 
    \log_\mathbf{o}(\mathbf{m}_i) = \mathbf{G}(\alpha_i)\log_\mathbf{o}(\mathbf{h}_i), \\ 
    \log_\mathbf{o}(\mathbf{t}_i) = \mathbf{G}(\beta_i)\log_\mathbf{o}(\mathbf{m}_i)\ \Rightarrow\ 
    \mathbf{G}(\theta_i) = \mathbf{G}(\alpha_i)\mathbf{G}(\beta_i)
\end{aligned}
\end{displaymath}
which holds true when $\theta_i=\alpha_i+\beta_i$ or $\theta_i=\alpha_i+\beta_i+2\pi$ or $\theta_i=\alpha_i+\beta_i-2\pi$.

\subsubsection{Proof for restricted rotation transformation}
\begin{theorem}
Restricted rotation transformation always satisfies the cone containment constraint.
\end{theorem}
\xproof
For any triple $(h, r, t)$ where $r$ is a hierarchical relation, we will prove that cone containment constraint is satisfied after the restricted rotation from $h$ to $t$, i.e., $\mathcal{C}_{f_2(\mathbf{h}_i, \mathbf{r}_i)}\subseteq\mathcal{C}_{\mathbf{h}_i}$. By the transitivity property of entailment cone as in Eq.~\ref{eq:transitivity}, we only need to prove $f_2(\mathbf{h}_i, \mathbf{r}_i)\in\mathcal{C}_{\mathbf{h}_i}$, which is
\begin{equation}
    \angle_{\mathbf{h}_i}f_2(\mathbf{h}_i, \mathbf{r}_i) \leq \phi_{\mathbf{h}_i}
\label{eq:proof}
\end{equation}
according to the cone expression in Eq.~\ref{eq:cone}.
We can calculate the angle, denoted as $\varphi$, on the left hand side of the equation in tangent space $\mathcal{T}_{\mathbf{h}_i}\mathcal{B}$ (which is equipped with Euclidean metric),
\begin{equation}
\begin{aligned}
    \varphi &= \angle_{\mathbf{h}_i}f_2(\mathbf{h}_i, \mathbf{r}_i) \\
    &= \angle(\log_{\mathbf{h}_i}(\frac{1+||\mathbf{h}_i||}{2||\mathbf{h}_i||}\mathbf{h}_i), \log_{\mathbf{h}_i}f_2(\mathbf{h}_i,\mathbf{r}_i)) \\
    &= \angle(\overline{\mathbf{h}}_i, \mathbf{G}(\theta_i\frac{\phi_\mathbf{\mathbf{h}_i}}{\pi})\overline{\mathbf{h}}_i) 
    = |\theta_i\frac{\phi_\mathbf{\mathbf{h}_i}}{\pi}|
\end{aligned}
\end{equation}
For $\theta_i\in[-\pi, \pi)$, we have $|\theta_i\frac{\phi_\mathbf{\mathbf{h}_i}}{\pi}|\leq \phi_{\mathbf{h}_i}$. Therefore Eq.~\ref{eq:proof} holds, suggesting that cone containment constraint is satisfied.
\subsection{Ablation studies on transformations in \name}
\label{sec:ablation}
Empirically, we show that our design of transformations in \name is effective: both restricted rotation transformation in the relation-specific subspace and the rotation transformation in the complement space are indispensable to the performance of our model on knowledge graph completion task.
\subsubsection{Ablation study on restricted rotation transformation}
\begin{table}[t]%
\centering  
\resizebox{\textwidth}{!}{
\begin{tabular}{c|ccc|ccc|ccc}
\hline
 & \multicolumn{3}{c}{All relations} & \multicolumn{3}{|c}{Hierarchical relations} & \multicolumn{3}{|c}{Non-hierarchical relations} \\
\hline
Model & MRR & H@1 & H@10 & MRR & H@1 & H@10 & MRR & H@1 & H@10\\
\hline
RotC  & .481 & .444 & .551 & .209 & .157 & .312 & .936 & .923 & .951 \\  
ConE & {\bf .496} & {\bf .453} & {\bf .579} & {\bf .231} & {\bf .171} & {\bf .355}  & {\bf .939} & {\bf .930} & {\bf .952} \\
\hline  
Improvement (\%) & +3.1\% & +2.0\% & +5.1\% & +10.5\% & +8.9\% & +13.8\% & +0.3\% & +0.7\% & +0.1\% \\
\hline
\end{tabular}
}
\vspace{1mm}
\caption{Results of ablation study on restricted rotation, for knowledge graph completion task on WN18RR. Results in three columns are conducted on all relations during evaluation, only hierarchical relations during evaluation and only non-hierarchical relations during evaluation.}
\label{tb:ablation}
\end{table}

\begin{table}[t]%
\centering
 \begin{tabular}{c|ccc} 
    \hline
    Relation & RotC & ConE & Improvement \\
    \hline
    \textit{hypernym} & .175 & {\bf .193} & +10.3\% \\
    \textit{instance hypernym} & .373 & {\bf .406} & +8.8\% \\
    \textit{member meronym} & .230 & {\bf .231} & +0.4\% \\
    \textit{synset domain topic of} & .382 & {\bf .413} & +8.1\% \\
    \textit{has part} & .208 & {\bf .213} & +2.4\% \\
    \textit{member of domain usage} & .200 & {\bf .345} & +72.5\% \\
    \textit{member of domain region} & .142 & {\bf .244} & +71.8\% \\
    \hline
\end{tabular}
\vspace{1mm}
\caption{Comparison of MRR for all hierarchical relations in WN18RR between RotC and ConE.}
\label{tb:each}
\end{table}

Restricted rotation transformation is vital in enforcing cone containment constraint, and thus it is indispensable to \name's performance on hierarchical tasks. However, its effect on knowledge graph completion task remains unknown. We further compare the performance of \name with one that does not use cone restricted rotation for modeling hierarchical relations, which we name as RotC. Specifically, RotC is the same as \name, except that it applies rotation transformation to all relations, and the cone angle loss as in Eq.~\ref{eq:angle} is excluded.

\xhdr{Results}
Ablation results are shown in Table~\ref{tb:ablation}. 
We can see remarkable improvement on knowledge graph completion task after applying restricted rotation transformation to hierarchical relations, especially in predicting missing hierarchical relations. 
The results suggest that restricted rotation transformation helps model hierarchical relation patterns.

\xhdr{Individual results for each hierarchical relation}
To further demonstrate that \name can deal with multiple hierarchical relations simultaneously with our proposed restricted rotation in subspaces, we report the improvement for knowledge graph completion on each type of missing hierarchical relation after adding cone restricted rotation, shown in Table~\ref{tb:each}. We observe significant improvement on \emph{all} hierarchical relations, which shows our way of modeling heterogeneous hierarchies to be effective.
Note that up to $72\%$ improvement is achieved for some hierarchical relation thanks to the restricted rotation operation in \name.

\begin{table}[t]%
\centering
 \begin{tabular}{c|cccc} 
    \hline
    Model & MRR & H@1 & H@3 & H@10 \\
    \hline
    ConE w/o rotation & .397 & .329 & .433 & .526 \\
    ConE & {\bf .496} & {\bf .453} & {\bf .515} & {\bf .579} \\
    \hline
\end{tabular}
\vspace{1mm}
\caption{Results of ablation study on rotation, for knowledge graph completion task on WN18RR. ConE w/o rotation is the model that applies restricted rotation in the whole embedding space for hierarchical relations.}
\vspace{-2mm}
\label{tb:ablation1}
\end{table}

\subsubsection{Ablation study on rotation transformation}
To address the importance of rotation transformation in modeling hierarchical relations, we present the performance comparison between ConE that uses rotation and one that does not use rotation for hierarchical relations on WN18RR.
The results in Table~\ref{tb:ablation1} suggest that rotation transformation for hierarchical relations is significant to the model’s expressive power.

\section{Strategies in assigning relation-specific subspace and embedding space curvature}
\label{sec:strategy}

We compare several strategies for assigning subspace for each hierarchical relation. 
For simplicity, we assign equal dimension subspaces for all hierarchical relations.

\subsection{Overlapping subspaces and orthogonal subspaces}

\begin{table}
\centering
\subtable[On knowledge graph completion]{
       \begin{tabular}{c|cccc} 
    \hline
    Model & MRR & H@1 & H@3 & H@10 \\
    \hline
    Orthogonal & .493 & .449 & .512 & .577 \\  
    Overlapping & {\bf .495} & {\bf .451} & {\bf .513} & {\bf .582} \\
    \hline
    \end{tabular}
}
\qquad
\subtable[On ancestor-descendant completion, in mAP metric]{        
      \begin{tabular}{c|ccc}
         \hline
         Model & 0\% & 50\% & 100\% \\
         \hline
        Orthogonal & {\bf .930} & {\bf .863} & .772 \\  
        Overlapping & .928 & .862 & {\bf .773} \\
        \hline
      \end{tabular}
}
\vspace{1mm}
\caption{Comparison between orthogonal subspaces and overlapping subspaces on WN18RR benchmark.}
\label{tb:subspace}
\vspace{-2mm}
\end{table}

\begin{figure}[t]
\centering
\subfigure[Performance on ancestor-descendant prediction (0\% inferred descendant pairs)]{
\begin{minipage}{0.45\linewidth}
\centering
\includegraphics[width=2.5in]{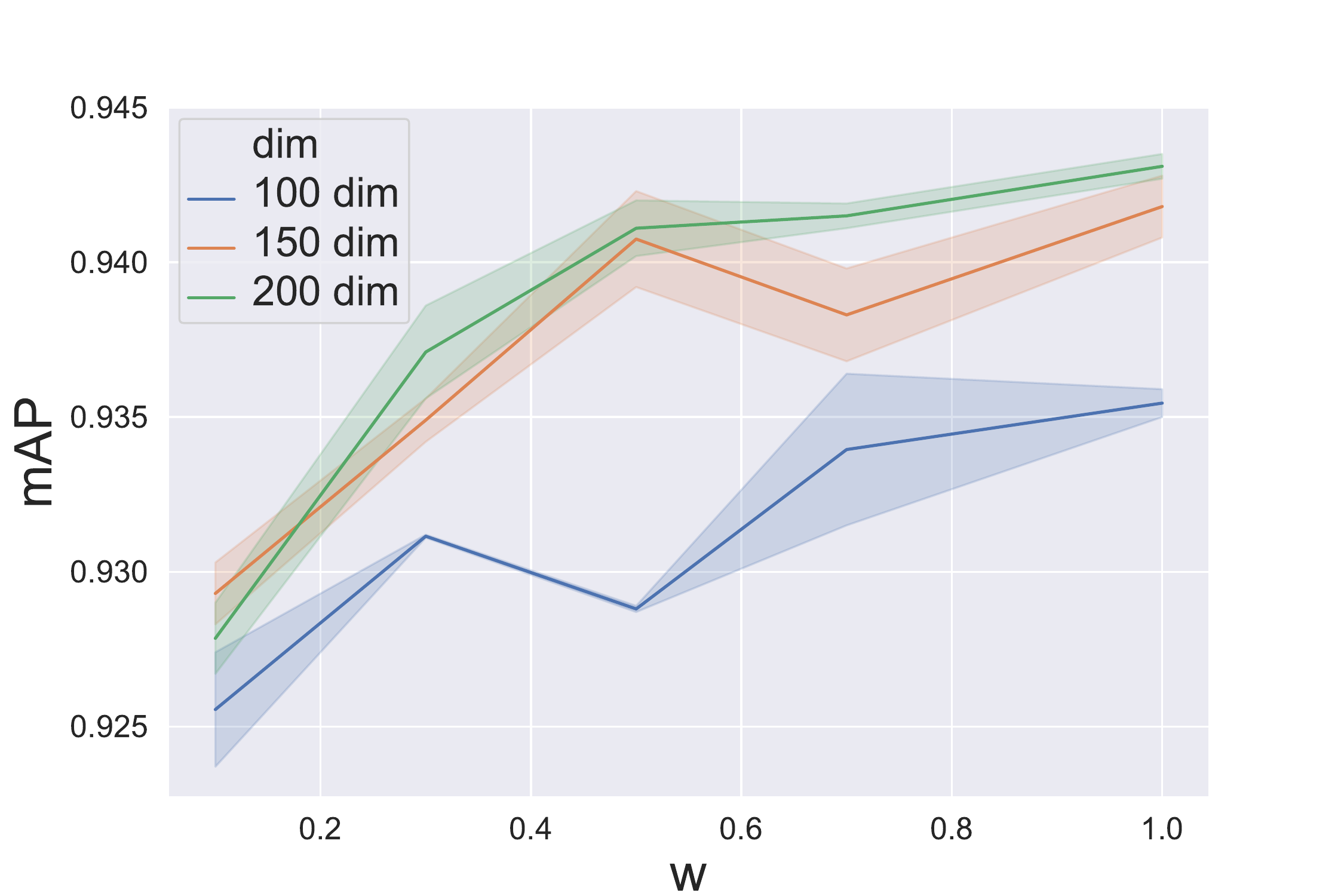}
\label{fig:strategy-map}
\end{minipage}%
}%
\qquad
\subfigure[Performance on knowledge graph completion task]{
\begin{minipage}{0.45\linewidth}
\centering
\includegraphics[width=2.5in]{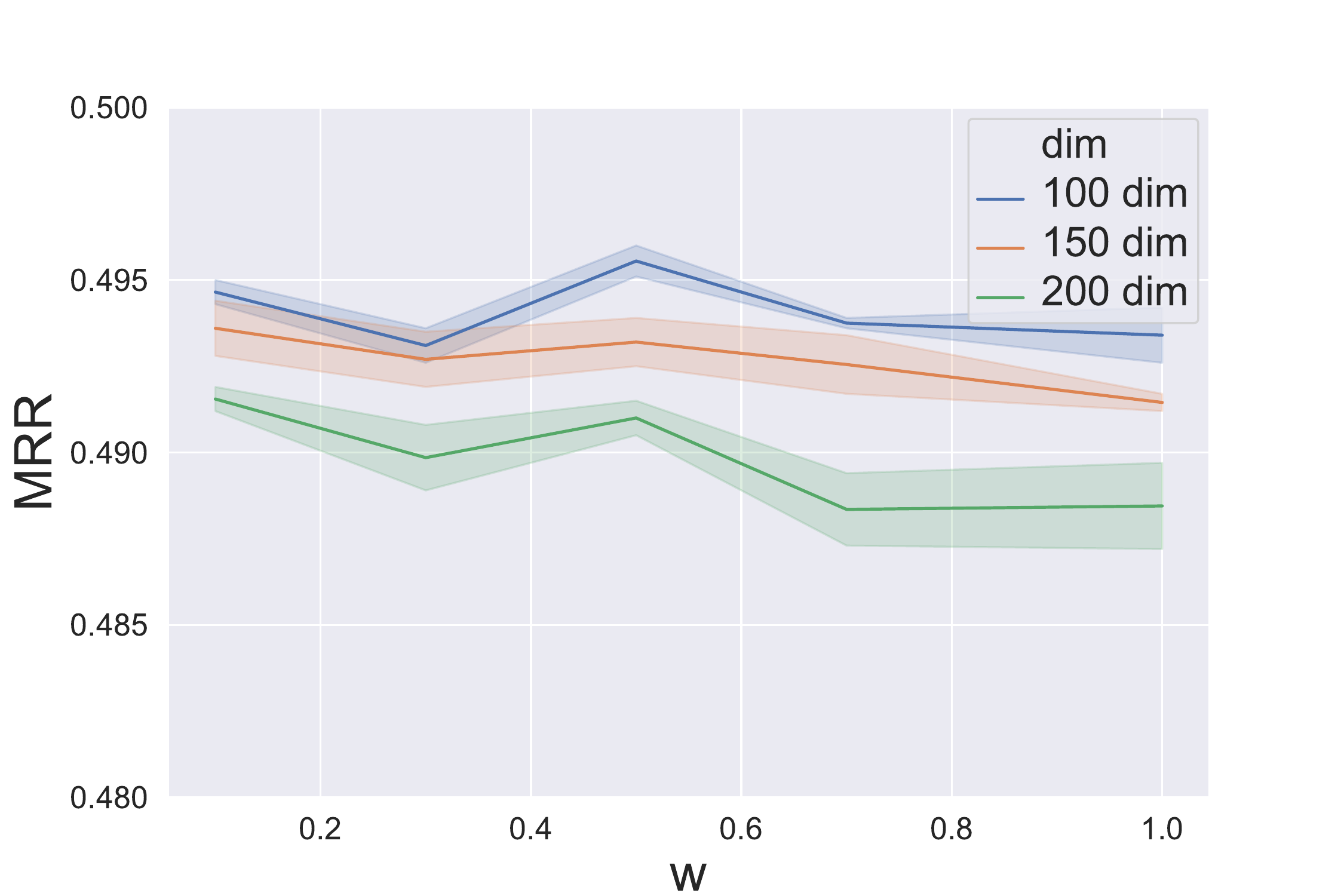}
\label{fig:strategy-mrr}
\end{minipage}%
}%
\centering
\caption{Performance of two tasks on WN18RR under varying strategies, including angle loss weight $w\in\{0.1, 0.3, 0.5, 0.7, 1.0\}$, dimension of subspace $d_s\in\{100,150,200\}$. Due to larger number of dimensions used per subspace, we use overlapping subspace strategy to assign relation-specific subspaces.}
\label{fig:strategy}
\vspace{-2mm}
\end{figure}

First, we compare the results on ancestor-descendant prediction and knowledge graph completion between different subspace assigning strategies, i.e., using overlapping subspaces and using orthogonal subspaces. 
We conduct the experiment on WN18RR dataset. For both strategies, the embedding dimension $d=500$ and the subspace dimension $d_s=70$ for each hierarchical relation (7 hierarchical relations in total hence it is possible to assign orthogonal subspaces). 
For assigning overlapping subspaces, since it is impossible to investigate all possible combinations, we randomly choose $d_s$ out of $d$ number of hyperbolic planes to each hierarchical relation.
To avoid the randomness of the results due to our method in assigning overlapping subspaces, we repeat the experiment multiple times and take the average for the final result.

\xhdr{Results}
Table~\ref{tb:subspace} reports the results on ancestor-descendant prediction task as well as knowledge graph completion task.
Between two strategies, \name performs slightly better on knowledge graph completion task under overlapping subspaces, while their performances are comparable on ancestor-descendant prediction task. 
The most significant advantage for using overlapping subspaces is that it does not suffer from limitation of subspace dimension, while for orthogonal subspaces the subspace dimension can be at most $d/n$ where $n$ is the number of hierarchical relations.

\subsection{Subspace dimension and angle loss weight}

We also study the effect of subspace dimension $d_s$ and angle loss weight $w$ (in Eq.~\ref{eq:total}) on the performance of \name. 
We use overlapping subspaces where we randomly choose $d_s$ out of $d=500$ hyperbolic planes to compose the subspace for each hierarchical relation.

\xhdr{Results}
Figure~\ref{fig:strategy} reports the results on both tasks in curves.
We notice a trade-off between two tasks for subspace dimension, where a larger dimension contributes to better performance on hierarchical task, while limiting the performance on knowledge graph completion task. 
With larger angle loss weight $w$, cone containment constraint is enforced more strictly, and thus the performance of \name on hierarchical task improves as shown in Figure~\ref{fig:strategy-map}. On the other hand, \name reaches peak performance on knowledge graph completion task at $w=0.5$.

\subsection{Space curvature}
Aside from setting fixed curvature $c=-1$, we also investigate on learning curvature, as \cite{chami2020low} suggests that fixing the curvature has a negative impact on performance of RotH.
With learning curvature, \name has (MRR, H@1, H@3, H@10) = (0.485, 0.441, 0.501, 0.570), on WN18RR benchmark, lower than original \name with fixed curvature with (MRR, H@1, H@3, H@10) = (0.496, 0.453, 0.515, 0.579).
The reason why RotH~\cite{chami2020low} needs learning space curvature while ConE does not lie in the choice of embedding space: RotH uses a $2d$-dimensional hyperbolic space while \name uses product space of $d$ hyperbolic planes. Our embedding space is less sensitive to its curvature, since for every subspace, the hierarchical structure for the corresponding single relation is less complex (than the entire hierarchy), and can thus be robust to choices of curvatures.

\section{Knowledge graph completion results in low dimensions}
\label{sec:low}
\begin{table}[t]%
\centering
 \begin{tabular}{c|cccc} 
    \hline
    Model & MRR & H@1 & H@3 & H@10 \\
    \hline
    RotatE & .387 & .330 & .417 & .491 \\  
    MuRP & .465 & .420 & .484 & \underline{.544} \\
    RotH & {\bf .472} & \underline{.428} & {\bf .490} & {\bf .553} \\
    \name & \underline{.471} & {\bf .436} & \underline{.486} & .537 \\
    \hline
\end{tabular}
\vspace{2mm}
\caption{Knowledge graph completion results for low-dimensional embeddings ($d=32$) on WN18RR. Best score in \textbf{bold} and second best \underline{underlined}.}
\label{tb:low}
\end{table}
One of the main benefits of learning embeddings in hyperbolic space is that it can model well even in low embedding dimensionalities. We report in Table~\ref{tb:low} the performance of \name in the low-dimensional setting for $d = 32$ on WN18RR dataset. Our performance is comparable to other hyperbolic embedding models (MuRP and RotH), while being superior to Euclidean embedding models (RotatE).

\section{Dataset details and GO21 dataset}
\label{sec:dataset_details}

WN18RR is a subset of WordNet \cite{miller1995wordnet}, which features lexical relationships between word senses. More than 60\% of all triples characterize hierarchical relationships. DDB14 is collected from Disease Database, which contains terminologies including diseases, drugs, and their relationships. Among all triples in DDB14, 30\% include hierarchical relations.

GO21 is a biological knowledge graph containing genes, proteins, drugs and diseases as entities, created based on several widely used biological databases, including Gene Ontology \cite{michael2000gene}, Disgenet \cite{pinero2019disgenet}, CTD \cite{davis2021ctd}, UMLS \cite{bodenreider2004umls}, DrugBank \cite{wishart2007drugbank}, ClassyFire \cite{feunang2016classyfire}, MeSH \cite{carolyn2019mesh} and PPI \cite{ruiz2020ppi}.
It contains 80k triples, while nearly 35\% of which include hierarchical relations.
The dataset will be made public at publication.

\section{AUROC results and hierarchy gap studies on ancestor-descendant prediction}
\label{sec:AUC}

\begin{table}[t]
\centering
\resizebox{\textwidth}{!}{
\begin{tabular}{c|ccc|ccc|ccc}
\hline 
 & \multicolumn{9}{c}{Percentage of inferred descendant pairs} \\
\hline
 & \multicolumn{3}{c|}{WN18RR} & \multicolumn{3}{c|}{DDB14} & \multicolumn{3}{c}{GO21} \\
\hline
Model & 0\% & 50\% & 100\% & 0\% & 50\% & 100\% & 0\% & 50\% & 100\% \\
\hline
Order & \underline{.859} & \underline{.676} & .495 & .971 & .745 & .533 & \underline{.643} & .587 & .542 \\
Poincar\'e & .784 & .649 & .511 & {\bf.981} & .763 & .541 & .534 & .529 & .526 \\
HypCone & .767 & .635 & .501 & .976 & \underline{.773} & \underline{.560} & .553 & .541 & .531 \\
RotatE & .599 & .601 & \underline{.599} & .506 & .505 & .514 & .622 & .607 & \underline{.598} \\
RotH & .559 & .567 & .574 & .525 & .519 & .517 & .630 & \underline{.612} & .594 \\
\hline
ConE & {\bf .874} & {\bf .762} & {\bf .657} & \underline{.979} & {\bf .888} & {\bf .802} & {\bf .704} & {\bf .653} & {\bf .601} \\
\hline
Improvement (\%) & +1.7\% & +12.7\% & +9.7\% & -0.2\% & +14.9\% & +43.2\% & +9.5\% & +6.7\% & +0.5\% \\
\hline
\end{tabular}
}
\vspace{1mm}
\caption{Ancestor-descendant prediction results in AUROC. Best score in \textbf{bold} and second best \underline{underlined}.}
\label{tb:auc}
\vspace{-2mm}
\end{table}

We show in Table~\ref{tb:auc} the results with AUROC (Area Under the Receiver Operating Characteristic curve) metric on ancestor-prediction tasks. 
It can be seen that the performance trend with AUROC metric is similar to that in Table~\ref{tb:map} with mAP metric.
\begin{figure}[t]
\centering
\subfigure[0\% inferred descendant pairs]{
\begin{minipage}{0.3\linewidth}
\centering
\includegraphics[width=1.8in]{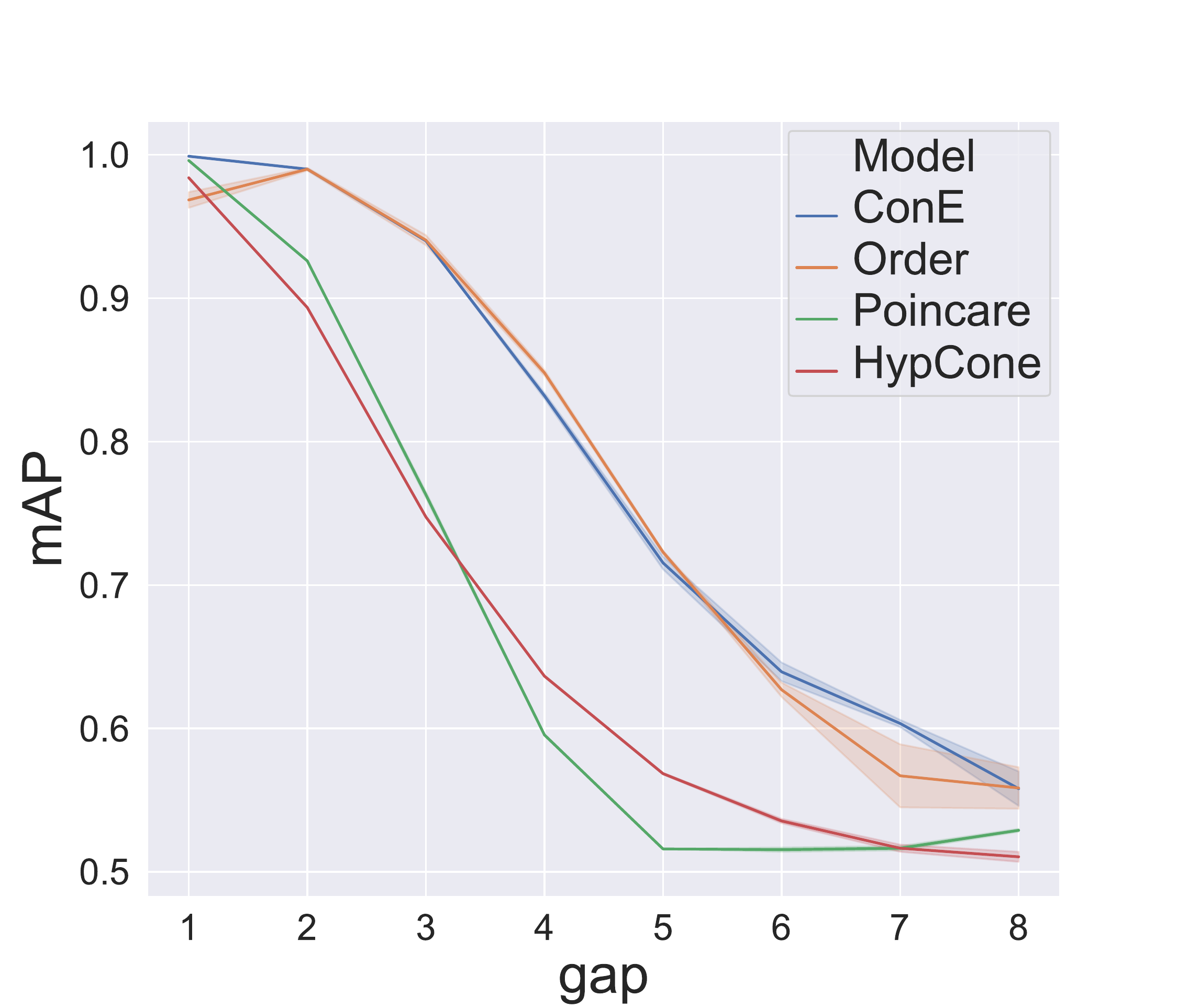}
\end{minipage}%
} \quad
\subfigure[50\% inferred descendant pairs]{
\begin{minipage}{0.3\linewidth}
\centering
\includegraphics[width=1.8in]{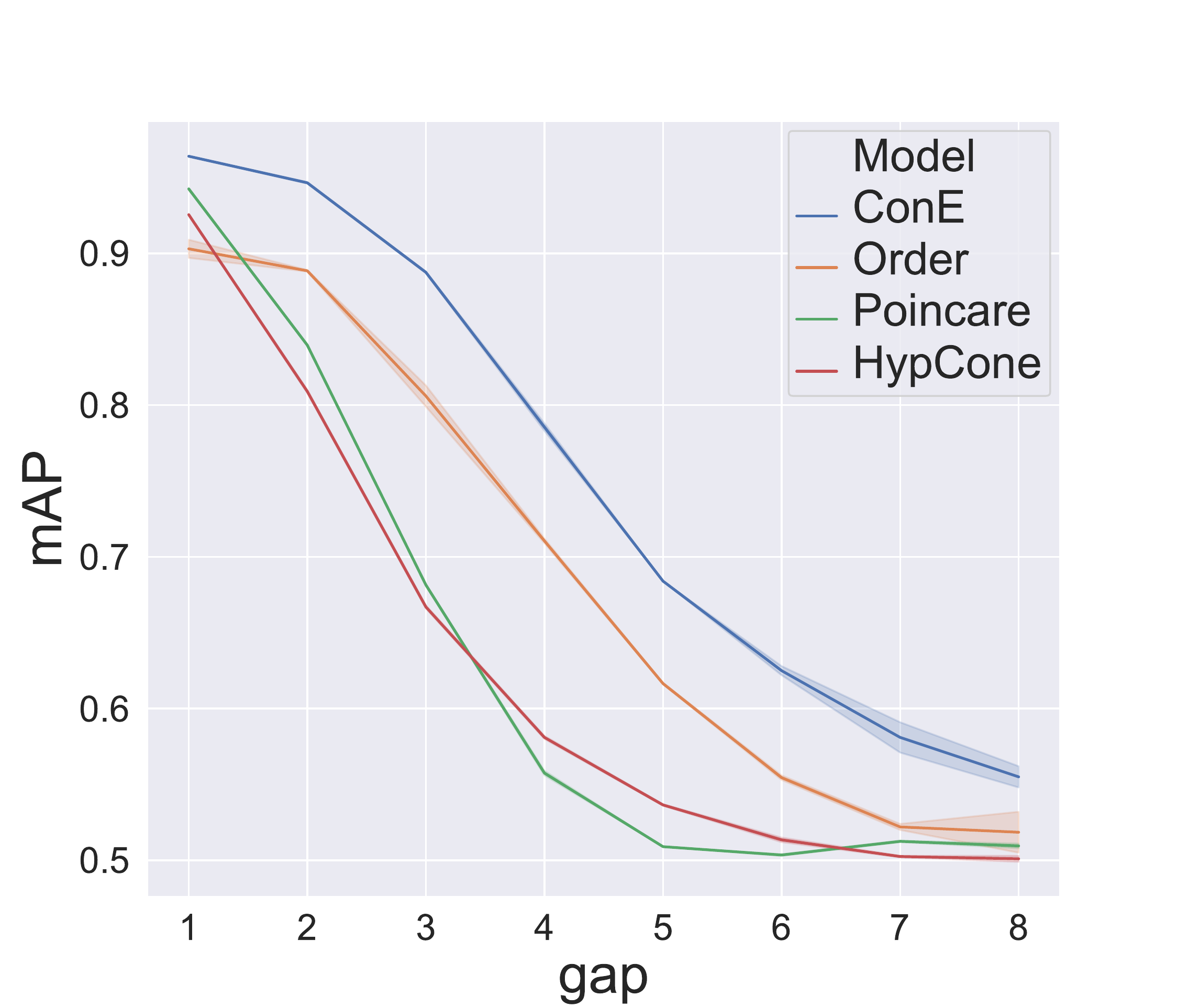}
\end{minipage}%
}\quad
\subfigure[100\% inferred descendant pairs]{
\begin{minipage}{0.3\linewidth}
\centering
\includegraphics[width=1.8in]{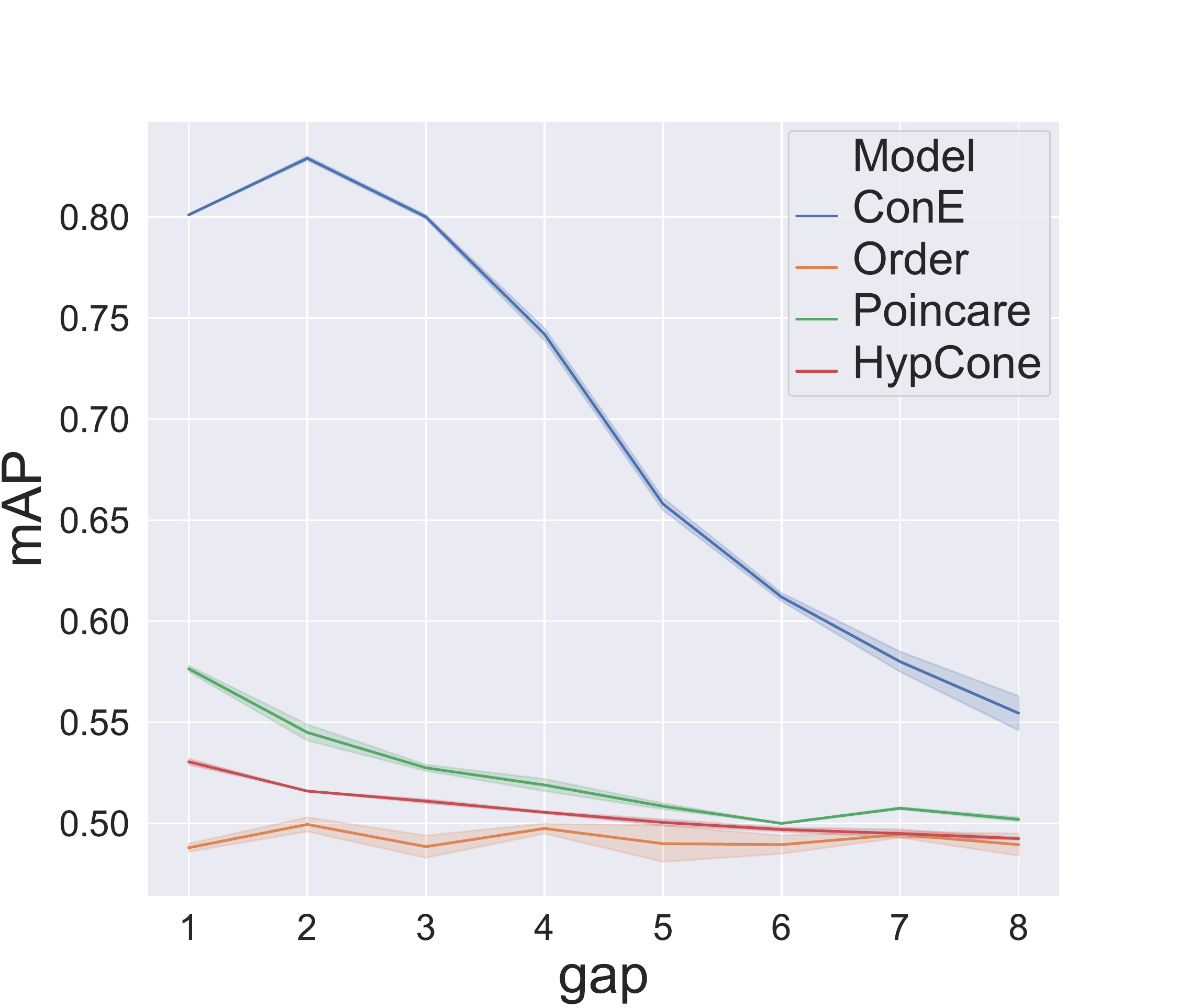}
\end{minipage}%
}%
\centering
\caption{mAP results on ancestor-descendant prediction under different hierarchy gaps (Def.~\ref{def:gap}) on WN18RR.}
\vspace{-2mm}
\label{fig:map}
\end{figure}
\begin{definition}{\textbf{Hierarchy gap.}}
The hierarchy gap of an ancestor-descendant pair $(u, v)$ is the length of path consisting of the same hierarchical relation connecting $u$ and $v$.
\label{def:gap}
\end{definition}
Moreover, we evaluate the classification performance of our model against other baselines over ancestor-descendant pairs with different hierarchy gaps (Def.~\ref{def:gap}), as shown in Figure~\ref{fig:map}. 
The trend of the curves is in line with our expectation: performance gets worse with larger hierarchy gaps.
Under the setting of 0\% inferred pairs, the performance of Poincar\'e embedding and Hyperbolic cone embedding drops dramatically as hierarchy gap increases, suggesting that transitivity is not well-preserved in these embeddings under heterogeneous setting.
In all settings (0\%, 50\% and 100\% inferred descendant pairs), \name significantly outperforms baselines.

\section{Hierarchical analysis: LCA prediction}
\label{sec:LCA}
We further demonstrate flexibility and power of \name using a new hierarchical task, lowest common ancestor (LCA) prediction.
Given two entities, we want to find the most distinguishable feature they have in common, e.g., \textit{LCA(WinePalm, SugarPalm)=PalmTree} in Figure~\ref{fig:hete_hier}. 
Formally, let $l_{uv}$ denote the hierarchy gap (Def.~\ref{def:gap}) between $u$ and $v$ and $l_{uv}=\infty$ if $u$ is not an ancestor of $v$, then we define $LCA(u, v) = \text{argmin}_{w\in\mathcal{E}}[(l_{wu}+l_{wv})]$.
Note that if multiple common ancestors have the same sum of hierarchy gap, we consider any of them to be correct.
\name uses ranking over all entities to predict LCA, with the following scoring function for $w$ to be the LCA of $u$ and $v$:
\begin{equation}
\begin{aligned}
    &\Phi_w(u, v) = \mathbf{m}\cdot(2\phi(\mathbf{w}_i)-\angle_{\mathbf{w}_i}\mathbf{u}_i-\angle_{\mathbf{w}_i}\mathbf{v}_i)_{i\in\{1,\cdots,d\}}
\end{aligned}
\end{equation}
We evaluate the LCA prediction task on WN18RR dataset, and use the embeddings of our trained \name model to rank and make prediction. Standard evaluation metrics including Hits at N (Hits@N) are calculated. Since no previous KG embedding method can directly perform the LCA task, we adapt them by training an MLP layer with the concatenation of the two entity embeddings as input and output the predicted entity (trained as a multi-label classification task).

\begin{table}[t]%
\centering  
\resizebox{\textwidth}{!}{
\begin{tabular}{c|ccc|ccc|ccc}
\hline
 & \multicolumn{3}{c|}{1-Hop} & \multicolumn{3}{c|}{2-Hop} & \multicolumn{3}{c}{3-Hop} \\
\hline
Model & H@1 & H@3 & H@10 & H@1 & H@3 & H@10 & H@1 & H@3 & H@10 \\
\hline 
Order \cite{vendrov2015order} & 39.2\% & 55.1\% & 61.6\% & 27.6\% & \underline{40.1\%} & \underline{54.7\%} & 15.1\% & 24.7\% & 42.2\% \\
Poincar\'e \cite{nickel2017poincare} & 1.5\% & 3.0\% & 8.0\% & \underline{31.4\%} & 34.6\% & 38.5\% & \underline{19.5\%} & 23.1\% & 38.5\% \\
HypCone \cite{ganea2018hyperbolic} & 15.0\% & 30.7\% & 53.0\% & 16.5\% & 30.1\% & 43.3\% & 12.4\% & \underline{38.1\%} & \underline{52.0\%} \\
RotatE \cite{sun2019rotate} & 54.7\% & 63.3\% & 69.7\% & 20.4\% & 29.0\% & 35.8\% & 14.6\% & 18.3\% & 20.7\% \\  
RotH \cite{chami2020low} & \underline{79.7\%} & \underline{86.0\%} & \underline{86.4\%} & 29.1\% & 35.7\% & 40.2\% & 13.9\% & 18.0\% & 21.9\% \\
\hline
ConE &  {\bf 98.1\%} & {\bf 99.3\%} & {\bf 99.4\%} & {\bf 48.6\%} & {\bf 89.6\%} & {\bf 97.3\%} & {\bf 24.2\%} & {\bf 55.6\%} & {\bf 80.6\%} \\
\hline  
\end{tabular}
}
\vspace{1mm}
\caption{LCA prediction task results on the WN18RR dataset. N-hop means that for any pair $(u, v)$ in the test set, the true LCA $w$ has hierarchy gaps (Def.~\ref{def:gap}) at most $N$ to $u$ and $v$. The task difficulty increases as the maximum number of hops to ancestor increases. Best score in \textbf{bold} and second best \underline{underlined}.}
\label{tb:LCA}
\end{table}

\xhdr{Results}
Table~\ref{tb:LCA} reports the LCA prediction results.  
\name can provide much more precise LCA prediction than baseline methods, and the performance gap increases as the number of hops to ancestor increases.
We summarize the reasons that \name performs superior to previous methods on LCA prediction: the task requires (1) the modeling of partial ordering for ancestor-descendant relation prediction and (2) an expressive embedding space for distinguishing the lowest ancestor. Only our \name model is able to do both.

\section{Training details}
\label{sec:training}

We report the best hyperparameters of \name on each dataset in Table~\ref{tb:hyperparameter}.
As suggested in \cite{ganea2018hyperbolic}, hyperbolic cone is hard to optimize with randomized initialization, so we utilize RotC model (which only involves rotation transformation) as pretraining for \name model, and recover the entity embedding from the pretrained RotC model with 0.5 factor.
For both the pretraining RotC model and \name model, we use Adam \cite{kingma2014adam} as the optimizer.
Self-adversarial training has been proven to be effective in \cite{sun2019rotate}, we also use self-adversarial technique during training for \name with self-adversarial temperature $\alpha=0.5$.

\xhdr{Knowledge graph completion} Standard evaluation metrics including Mean Reciprocal Rank (MRR), Hits at N (H@N) are calculated in the filtered setting where all true triples are filtered out during ranking.

In our experiments, we train and evaluate our model on a single GeForce RTX 3090 GPU.
We train the model for 500 epochs, 1000 epochs, 100 epochs and 600 epochs on WN18RR, DDB14, GO21 and FB15k-237 respectively, and the training procedure takes 4hrs, 2hrs, 6hrs, 6hrs on these four datasets.
On knowledge graph completion task, \name model has standard deviation less than 0.001 on MRR metric across all datasets. On ancestor-descendant classification task, \name model has standard deviation less than 0.01 on mAP metric across all datasets.

For all baselines mentioned in our work, we also perform comprehensive hyperparameter search. Specifically, for KG embedding methods (TransE~\cite{bordes2013translating}, RotatE~\cite{sun2019rotate}, TuckER~\cite{balavzevic2019tucker}, HAKE~\cite{zhang2020learning}, MuRP~\cite{balazevic2019multi}, RotH~\cite{chami2020low}), we search for embedding dimension in $\{100, 250, 500\}$, batch size in $\{256, 512, 1024\}$, learning rate in $\{0.01, 0.001, 0.0001\}$ and negative sampling size in $\{50, 100, 250\}$.
For partial order modeling methods (Order~\cite{vendrov2015order}, Poincar\'e~\cite{nickel2017poincare}, HypCone~\cite{ganea2018hyperbolic}), we search for embedding dimension in $\{50, 100, 250, 500\}$ and learning rate in $\{0.001, 0.0001, 0.00001\}$.
\begin{table}[t]%
\centering  
\resizebox{\textwidth}{!}{
\begin{tabular}{c|cccccc}
\hline
Dataset & embedding dim & learning rate & batch size & negative samples & subspace dim & angle loss weight \\
\hline
WN18RR & 500 & 0.001 & 1024 & 50 & 100 & 0.5 \\  
DDB14 & 500 & 0.001 & 1024 & 50 & 50 & 0.7\\
GO21 & 500 & 0.005 & 1024 & 50 & 50 & 0.1\\
FB15k-237 & 500 & 0.0001 & 1024 & 100 & - & - \\
\hline
\end{tabular}
}
\vspace{1mm}
\caption{Best hyperparameter setting of \name on four datasets.}
\vspace{-2mm}
\label{tb:hyperparameter}
\end{table}

\section{Krackhardt hierarchical measurement}
\label{krackhardt}
\subsection{Krackhardt score for the whole graph}
The paper \cite{krackhardt1994graph} proposes a set of scores to measure how hierarchical a graph is. It includes four scores: \emph{(connectedness, hierarchy, efficiency, LUBedness)}. Each score range from 0 to 1, and higher scores mean more hierarchical. When all four scores equal to 1, the digraph is a tree, normally considered as the most hierarchical structure. We make some adjustments to the computation of the metrics from the original paper to adapt them to heterogeneous graphs.

\emph{1. Connectedness}.
Connectedness measures the connectivity of a graph, where a connected digraph (each node can reach every other node in the underlying graph) will be given score 1 and the score goes down with more disconnected pairs. Formally, the degree of connectedness is
\begin{equation}
    \emph{connectedness} = \frac{c}{n(n-1)/2}
\end{equation}
where $c$ is the number of connected pairs and $n$ is the total number of nodes.

\emph{2. Hierarchy}. 
Hierarchy measures the order property of the relations in the graph. If for each pair of nodes such that one node $u$ can reach the other node $v$, $v$ cannot reach $u$, then the hierarchy score is 1. In knowledge graph this implies that if $\textit{(u, rel, v)}\in \mathbb{T}$ then $\textit{(v, rel, u)}\notin \mathbb{T}$. Let $T$ denote the set of ordered pairs (u, v) such that $u$ can reach $v$, and $S=\{(v, u)|(u, v)\in T, v\ cannot\ reach\ u\}$, the degree of hierarchy is defined as
\begin{equation}
    \emph{hierarchy} = \frac{|S|}{|T|}
\end{equation}

\emph{3. Efficiency}. 
Another condition to make sure that a structure is a tree is that the graph contains exactly $n-1$ edges, given $n$ number of nodes. In other word, the graph cannot have redundant edges. The degree of efficiency is defined as
\begin{equation}
    \emph{efficiency} = 1-\alpha\cdot\frac{m-(n-1)}{(n-1)(n-2)/2}
\end{equation}
where $m$ is the number of edges in the graph. Numerator $m-(n-1)$ is the number of redundant edges in the graph while denominator $(n-1)(n-2)/2$ is the maximum number of redundant edges possible. In the original paper \cite{krackhardt1994graph}, $\alpha$ is set to 1, in our case we take $\alpha=500$ to make the gap larger since common knowledge graph are always sparse.

\emph{4. LUBedness}. 
The last condition for a tree structure is that every pair of nodes has a least upper bound, which is the same as our defined LCA concept (in Sec.~\ref{sec:LCA}) in knowledge graph case. Different from the homogeneous setting in \cite{krackhardt1994graph}, we still restrict LCA to a single relation (same relation on the paths between the pair of nodes and their LCA), since heterogeneous hierarchies may exist in a single KG. Let $T=\{(u, v)|(u, v)\ has\ a\ LCA\}$, then the degree of LUBedness is defined as
\begin{equation}
    \emph{LUBedness} = \frac{|T|}{n(n-1)}
\end{equation}

\subsection{Hierarchical-ness scores for each relation}
Here we introduce the Hierarchical-ness scores for each relation, which is a modified version of original Krackhardt scores on the induced subgraph of a relation.
We observe, using the groundtruth hypernym, hyponym and non-hierarchical relations in existing datasets (WN18RR, DDB14, GO21), that the Hierarchical-ness scores for hypernym, hyponym and non-hierarchical relations can be easily separated via decision boundaries.
To apply \name on a dataset where the type of relation is not available, we can compute the Hierarchical-ness scores of the relations, and classify the hierarchical-ness of the relations via the decision boundaries.

Here we introduce the computation of our Hierarchical-ness scores, which contain two terms: \emph{(asymmetry, tree\c likeness)}.

\emph{1. Asymmetry}. 
The \emph{asymmetry} metric is the same as \emph{hierarchy} metric in Krackhardt scores. 

\emph{2. Tree\c likeness}. 
The \emph{tree\c likeness} metric is adapted from the \emph{LUBedness} metric in Krackhardt scores where three adjustments are made: 

\textbf{(a)} The subgraph induced by a single relation is not guaranteed to be connected, and forest is a typical hierarchical structure in such a disconnected graph. We cannot make sure every pair of nodes are in the same tree, and thus we evaluate on all connected pairs and check whether they have an LCA.
Let $P$ denote the set of pairs $(u, v)$ such that $u$ and $v$ are connected, and the set $Q=\{(u, v)|(u, v)\in P\ and\ (u, v)\ has\ a\ LCA\}$. Then our new \emph{LUBedness'} for disconnected graph is calculated as
\begin{equation}
    \emph{LUBedness'} = \frac{|Q|}{|P|}
\end{equation}

\textbf{(b)} We want to distinguish true hierarchical relations from common 1-N relations, where the transitivity property may not hold (for example, \emph{participants} of some event entity is a 1-N relation, yet it does not define a partial ordering since the head entity and tail entity are not the same type of entities). This kind of relation can be characterized by 1-depth trees in their induced subgraph, while hierarchical relations usually induce trees of greater depth. Hence we add punishment to the induced subgraphs containing mostly 1-depth trees to exclude non-hierarchical 1-N relations. In particular, let $E$ denote the set of edges, and $S=\{u|\exists v: (u, v)\in E\ or\ (v, u)\in E\}$, $T=\{u|\exists v: (u, v)\in E\ and\ (v, u)\in E\}$. If 1-depth trees are prevalent in the structure, then $|T|$ is approximately 0. We define the punishment decaying factor (lower means more punishment):
\begin{equation}
    \emph{d} = \frac{|T|}{|S|}
\end{equation}

\textbf{(c)} \emph{LUBedness} metric also depends on the direction of the relation, since LCA exists only if the relations are hyponym type (pointing from parent node to child nodes) while hypernym type relation can also define a partial ordering and considered as hierarchical relation. Hence for each relation, we define two induced graphs $G$ and $G_{rev}$, $G$ in original direction and $G_{rev}$ in reversed direction. We calculate the \emph{LUBedness} metric of the two graphs, if the score of $G$ is much higher than the score of $G_{rev}$ then the relation is of hyponym type, and vice versa. We take the absolute value of $LUBedness'(G)-LUBedness'(G_{rev})$ as the score to measure the hierarchical-ness while its sign to check if it is of hypernym type or hyponym type. 

Finally, our \emph{tree\c likeness} metric is calculated through
\begin{equation}
    \emph{tree\c likeness} = (LUBedness'(G)-LUBedness'(G_{rev})) / \max(1, (\log_{10}(d))^2)
\end{equation}

\begin{figure}[t]
\centering
\subfigure[Hierarchical-ness score visualization for all relations.]{
\begin{minipage}{0.5\linewidth}
\centering
\includegraphics[width=0.8\linewidth]{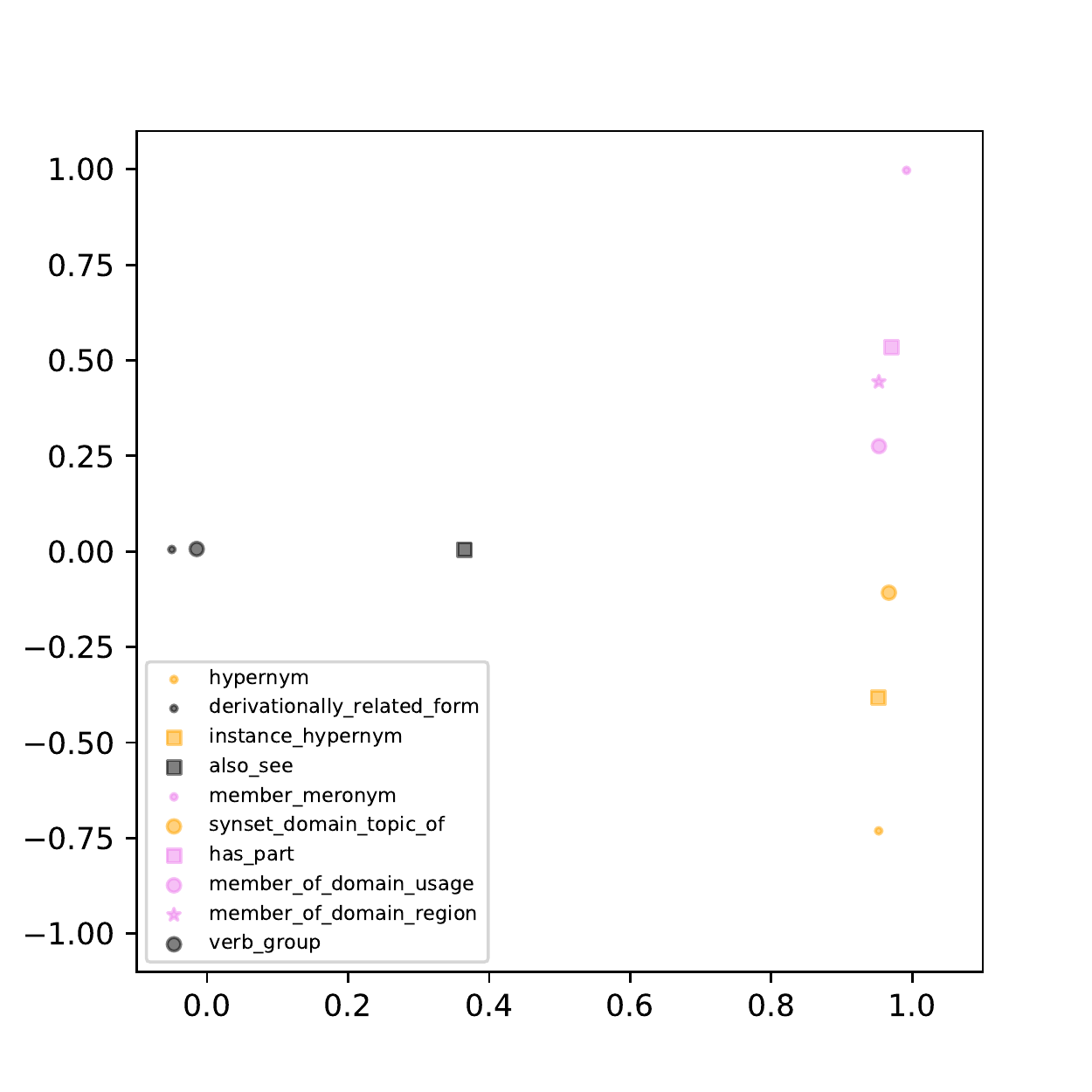}
\label{fig:wn18rr_vis}
\end{minipage}%
}
\qquad
\subfigure[Relation type classification based on Hierarchical-ness score.]{
\begin{minipage}{0.4\linewidth}
\centering
\vspace{13mm}
\resizebox{1\textwidth}{!}{
\begin{tabular}{c|c|c}
   Relation  & Score & Hierarchical \\
   \hline
   \textit{hypernym}  &  2.0 & true \\
   \textit{member meronym} & 2.0 & true \\
   \textit{has part} & 2.0 & true \\
   \textit{member of domain region} & 1.9 & true \\
   \textit{instance hypernym} & 1.7 & true \\
   \textit{member of domain usage} & 1.5 & true \\
   \textit{synset domain topic of} & 1.2 & true \\
   \hline
   \textit{also see} & 0.4 & false \\
   \textit{derivationally related form} & 0.0 & false \\
   \textit{verb group} & 0.0 & false \\
   \textit{similar to} & 0.0 & false \\
   \multicolumn{3}{c}{\vspace{10mm}}
\end{tabular}
}
\label{fig:wn18rr_score}
\end{minipage}%
}
\label{fig:wn18rr}
\caption{Results on WN18RR dataset. (a) Score \emph{(asymmetry, tree\c likeness)} as $(x, y)$ coordinate in visualization. Orange dots denote hypernym type relations, violet dots denote hyponym type relations and black dots denote non-hierarchical relations. (b) Relations above the line are predicted to be hierarchical relations, and ground-truth relation type are in the third column. All relations are correctly predicted.}
\end{figure}

\begin{figure}[t]
\centering
\subfigure[Hierarchical-ness score visualization for all relations.]{
\begin{minipage}{0.5\linewidth}
\centering
\includegraphics[width=0.8\linewidth]{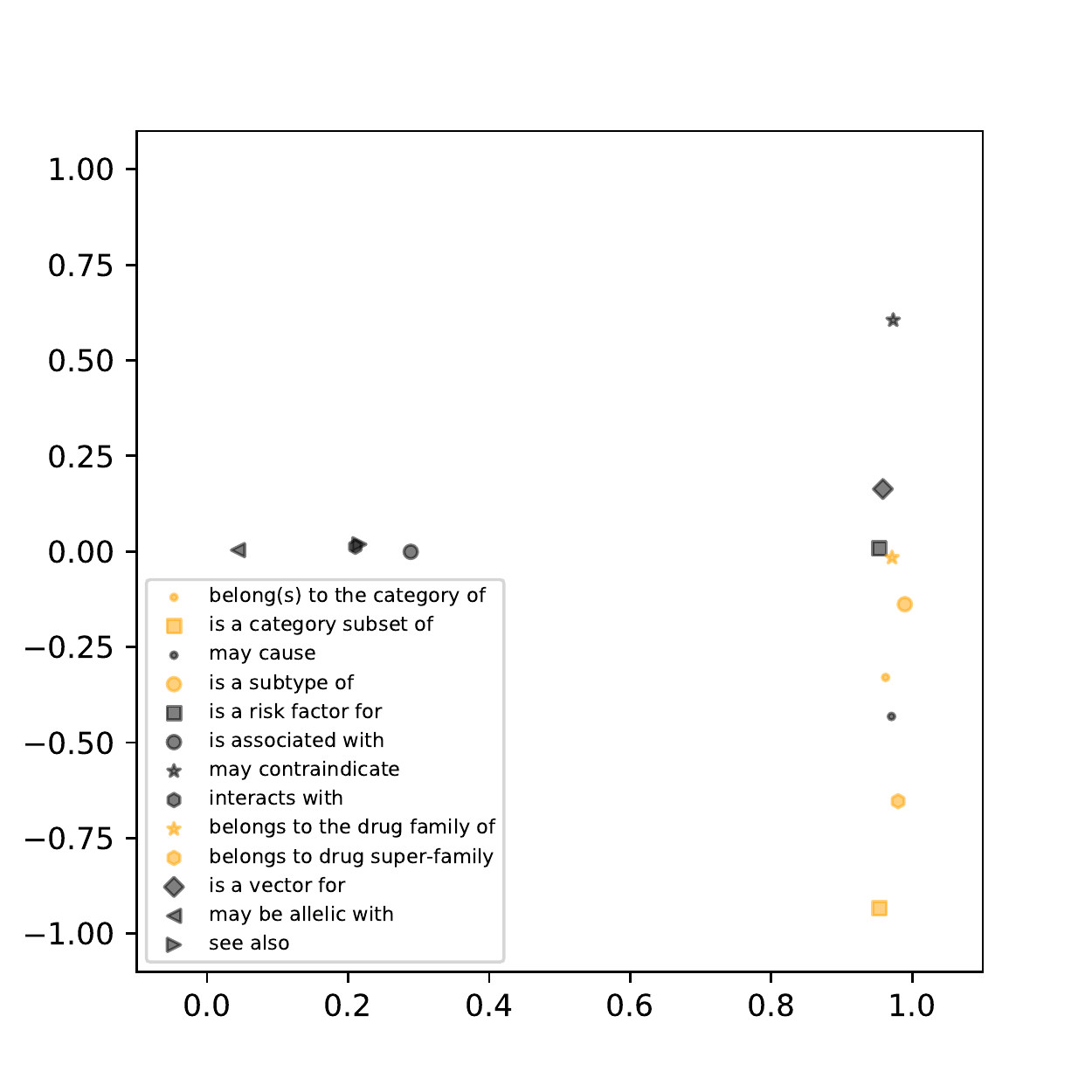}
\label{fig:DDB14_vis}
\end{minipage}%
}
\qquad
\subfigure[Relation type classification based on Hierarchical-ness score.]{
\begin{minipage}{0.4\linewidth}
\centering
\vspace{8mm}
\resizebox{1\textwidth}{!}{
\begin{tabular}{c|c|c}
   Relation  & Score & Hierarchical \\
   \hline
   \textit{is a category subset of}  &  2.0 & true \\
   \textit{belongs to drug super-family} & 2.0 & true \\
   \textit{has part} & 2.0 & true \\
   \textit{may contraindicate} & 1.9 & false$^*$ \\
   \textit{may cause} & 1.8 & false$^*$ \\
   \textit{belong(s) to the category of} & 1.7 & true \\
   \textit{belong(s) to the category of} & 1.2 & true \\
   \textit{is a vector for} & 1.3 & false$^*$ \\
   \textit{is a subtype of} & 1.3 & true \\
   \textit{belongs to the drug family of} & 1.1 & true \\
   \hline
   \textit{is a risk factor for} & 1.0 & false \\
   \textit{see also} & 0.3 & false \\
   \textit{interacts with} & 0.3 & false \\
   \textit{may be allelic with} & 0.1 & false \\
   \textit{is an ingredient of} & 0.0 & false \\
   \multicolumn{3}{c}{\vspace{3mm}}
\end{tabular}
}
\label{fig:DDB14_score}
\end{minipage}%
}
\label{fig:DDB14}
\caption{Results on DDB14 dataset. (a) Score \emph{(asymmetry, tree\c likeness)} as $(x, y)$ coordinate in visualization. Orange dots denote hypernym type relations, violet dots denote hyponym type relations and black dots denote non-hierarchical relations. (b) Relations above the line are predicted to be hierarchical relations, and ground-truth relation type are in the third column. Predictions are correct except three non-hierarchical relations are inferred to be hierarchical relations, while these relations do have soft-hierarchical property.}
\end{figure}

\begin{figure}[t]
\centering
\subfigure[Hierarchical-ness score visualization for all relations.]{
\begin{minipage}{0.5\linewidth}
\centering
\includegraphics[width=0.8\linewidth]{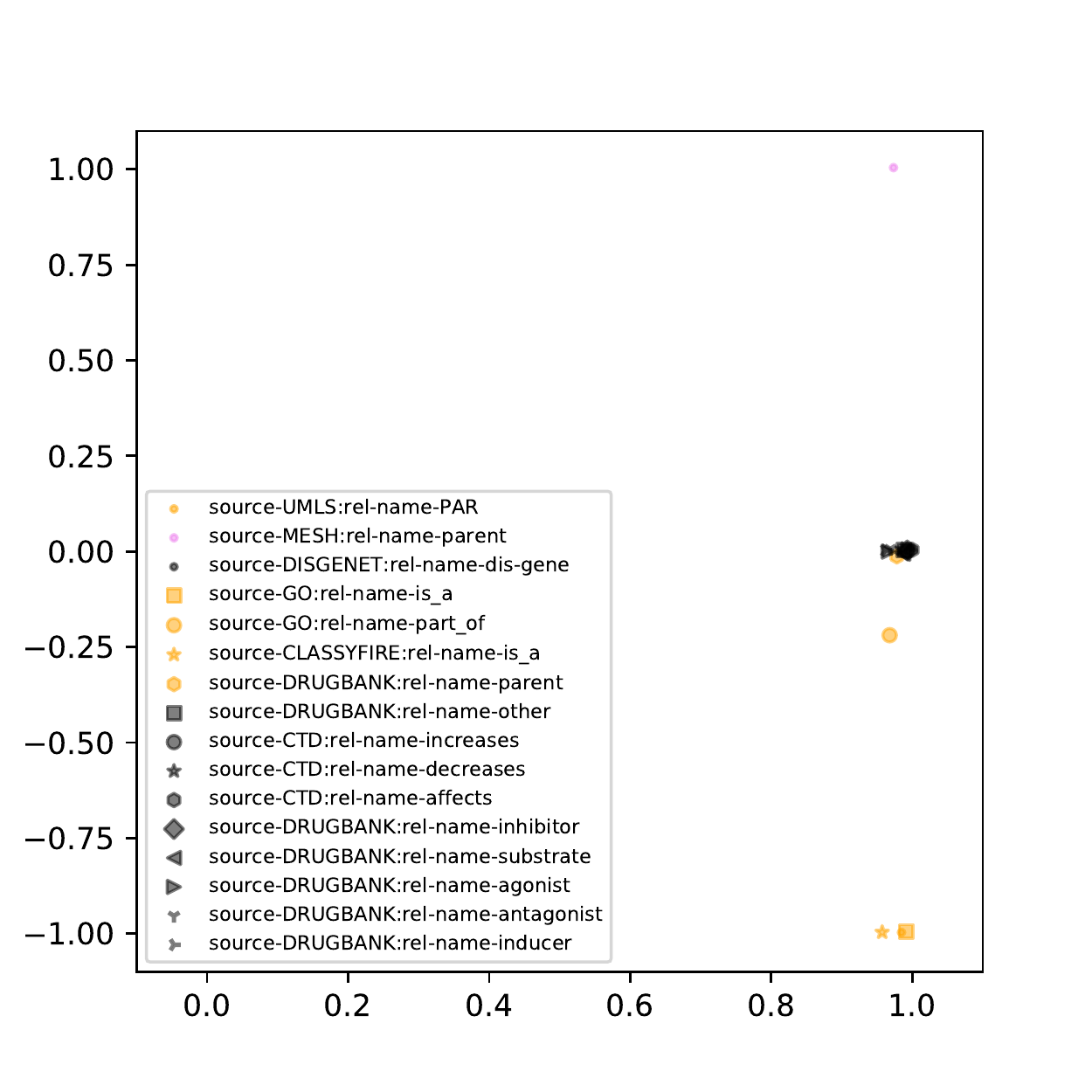}
\label{fig:GO21_vis}
\end{minipage}%
}
\qquad
\subfigure[Relation type classification based on Hierarchical-ness score.]{
\begin{minipage}{0.4\linewidth}
\centering
\vspace{13mm}
\resizebox{1\textwidth}{!}{
\begin{tabular}{c|c|c}
   Relation  & Score & Hierarchical \\
   \hline
   \textit{source-GO:rel-name-is\c a}  & 2.0 & true \\
   \textit{source-MESH:rel-name-parent} & 2.0 & true \\
   \textit{source-CLASSYFIRE:rel-name-is\c a} & 2.0 & true \\
   \textit{source-UMLS:rel-name-PAR} & 2.0 & true \\
   \textit{source-GO:rel-name-part\c of} & 1.2 & true \\
   \hline
   \textit{rest of the relations} & $<1.1$ & false \\
   \multicolumn{3}{c}{\vspace{10mm}}
\end{tabular}
}
\label{fig:GO21_score}
\end{minipage}%
}
\label{fig:GO21}
\caption{Results on GO21 dataset. (a) Score \emph{(asymmetry, tree\c likeness)} as $(x, y)$ coordinate in visualization. Orange dots denote hypernym type relations, violet dots denote hyponym type relations and black dots denote non-hierarchical relations. (b) Relations above the line are predicted to be hierarchical relations, and ground-truth relation type are in the third column. All relations are correctly predicted.}
\end{figure}

\begin{figure}[t]
\centering
\subfigure[Hierarchical-ness score visualization for all relations.]{
\begin{minipage}{0.5\linewidth}
\centering
\includegraphics[width=0.8\linewidth]{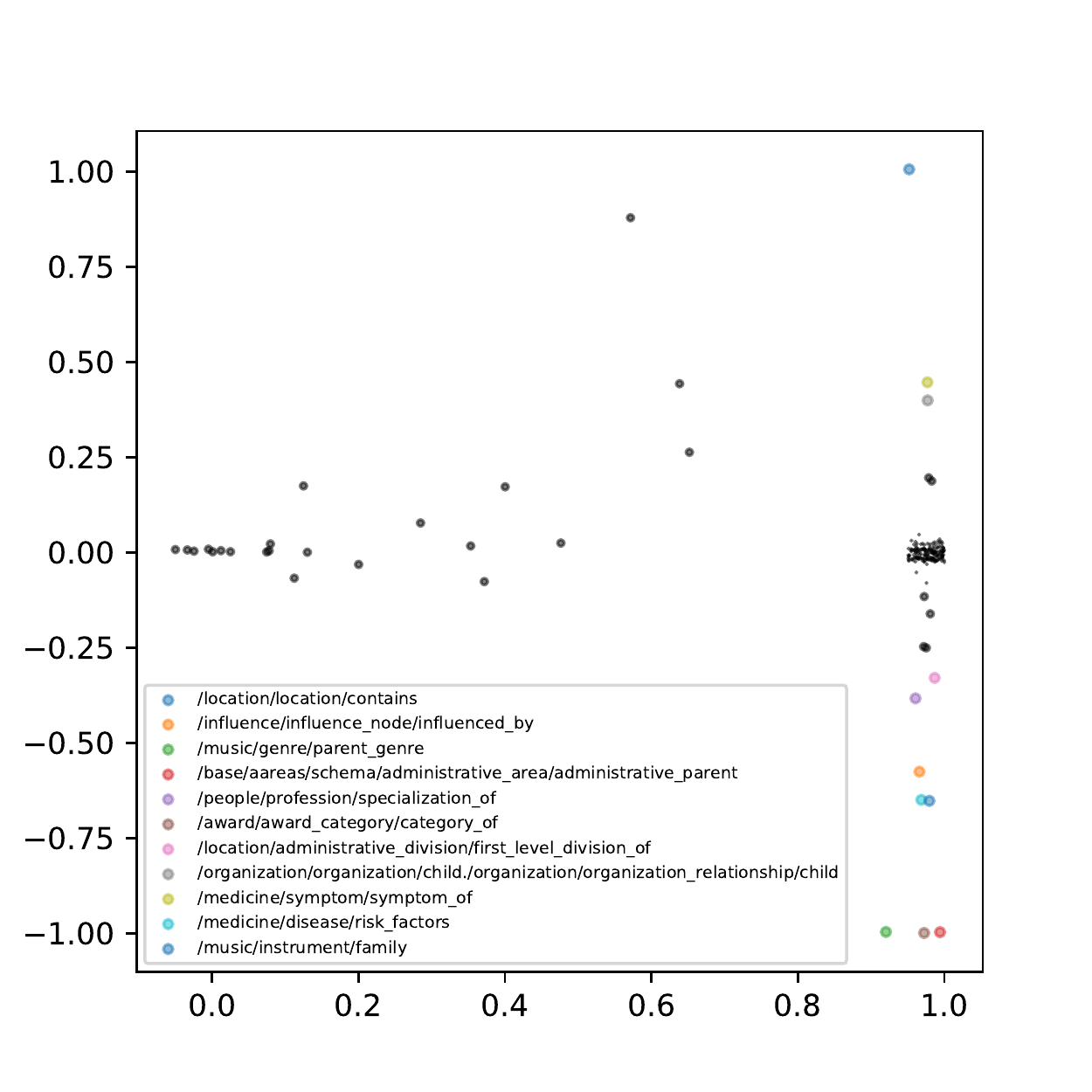}
\label{fig:FB15K237_vis}
\end{minipage}%
}
\qquad
\subfigure[Relation type classification based on Hierarchical-ness score.]{
\begin{minipage}{0.4\linewidth}
\centering
\vspace{13mm}
\resizebox{1\textwidth}{!}{
\begin{tabular}{c|c|c}
   Relation  & Score & Hierarchical \\
   \hline
   \textit{administrative\c area/administrative\c parent}  & 2.0 & true \\
   \textit{award\c category/category\c of} & 2.0 & true \\
   \textit{music/genre/parent\c genre} & 2.0 & true \\
   \textit{location/contains} & 2.0 & true \\
   \textit{music/instrument/family} & 1.7 & true \\
   \textit{medicine/disease/risk\c factors} & 1.7 & unknown \\
   \textit{influence/influence\c node/influenced\c by} & 1.6 & unknown \\
   \textit{medicine/symptom/symptom\c of} & 1.4 & unknown \\
   \textit{organization\c relationship/child} & 1.4 & true \\
   \textit{administrative\c division/first\c level\c division\c of} & 1.3 & true \\
   \hline
   \textit{rest of the relations} & $<1.1$ & false \\
   \multicolumn{3}{c}{\vspace{10mm}}
\end{tabular}
}
\label{fig:FB15K237_score}
\end{minipage}%
}
\label{fig:FB15K237}
\caption{Results on FB15k-237 dataset. (a) Score \emph{(asymmetry, tree\c likeness)} as $(x, y)$ coordinate in visualization. Dots with non-black colors denote top hierarchical-like relations among all 237 relations (their meanings are annotated in lower left of the figure). (b) Relations above the line are predicted to be hierarchical relations, and ground-truth relation type in column 3 are manually labeled.}
\end{figure}

We show that our Hierarchical-ness scores indicate the type of relation on WN18RR, DDB14 and GO21 datasets. 
In Figure~\ref{fig:wn18rr_vis}, Figure~\ref{fig:DDB14_vis}, Figure~\ref{fig:GO21_vis}, we visualize the two-dimensional scores for each relation in the three datasets. 
Groundtruth hypernym type relations are colored in orange and hyponym relations are colored in violet. 
We can see that hypernym type relations are clustered in lower right while hyponym type relations are clustered in upper right, indicating that hyponym type relations have large \emph{asymmetry} score and large positive \emph{tree\c likeness} score while hyponym type relations have large \emph{asymmetry} score and large negative \emph{tree\c likeness} score.
Moreover, We use $\emph{asymmetry}+|\emph{tree\c likeness}|$ as the total Hierarchical-ness score and set threshold 1.1 to separate hierarchical relations and non-hierarchical relations.
As shown in Figure~\ref{fig:wn18rr_score}, Figure~\ref{fig:DDB14_score}, Figure~\ref{fig:GO21_score}, our classification results highly conform to the groundtruth relation type.

Additionally, we use our Hierarchical-ness scores to distinguish hierarchical relations from 237 relations in FB15k-237, as shown in Figure~\ref{fig:FB15K237_vis}, Figure~\ref{fig:FB15K237_score}.
Since there is no labeling of relation type in FB15k-237, we do not have groundtruth.
We label the relations that rank highest on Hierarchical-ness score and discover that they are indeed hierarchical relations (suggested by keywords in their name, such as ``child'', ``parent'').

\end{document}